\documentclass[journal,twoside]{IEEEtran}

\usepackage{times}
\usepackage{graphicx,color}
\usepackage{algorithmic}
\usepackage{amsfonts,amssymb,amsmath,xspace}
\usepackage{mathrsfs,subfigure}

\usepackage{amsthm}
\theoremstyle{plain}
\newtheorem{lemma}{Lemma}
\newtheorem{theorem}{Theorem}

\newtheorem{corollary}{Corollary}

\newcommand{\Pd}{P_{\textrm{d}}}

\newcommand{\lfa}{\lambda_{\textrm{fa}}}
\newcommand{\lfan}{\lambda_{\textrm{fan}}}
\newcommand{\lambdan}{\lambda_{\textrm{new}}}

\newcommand{\nuv}{\vec{\nu}}

\renewcommand{\eqref}[1]{(\ref{#1})}
\newcommand{\eqsref}[1]{Eqs.~(\ref{#1})}
\newcommand{\noeqref}[1]{(\ref{#1})}

\newcommand{\figref}[1]{Fig.~\ref{#1}}

\newcommand{\secref}[1]{Section~\ref{#1}}

\newcommand{\lemref}[1]{Lemma~\ref{#1}}

\newcommand{\thref}[1]{Theorem~\ref{#1}}

\newcommand{\coref}[1]{Corollary~\ref{#1}}

\newcommand{\apost}{\textit{a posteriori}\xspace}
\newcommand{\ie}{\textit{i.e.}\xspace}
\newcommand{\eg}{\textit{e.g.}\xspace}

\newcommand{\etc}{\textit{etc}\xspace}

\newcommand{\dif}{\mathrm{d}}

\newcommand{\calE}{{\cal E}}

\newcommand{\calG}{{\cal G}}

\newcommand{\calM}{{\cal M}}
\newcommand{\calN}{{\cal N}}

\newcommand{\calT}{{\cal T}}

\newcommand{\calX}{{\cal X}}

\newcommand{\Rbb}{\mathbb{R}}

\renewcommand{\vec}[1]{\boldsymbol{#1}}

\newcommand{\gv}{\vec{g}}
\newcommand{\hv}{\vec{h}}

\newcommand{\muv}{\vec{\mu}}

\title{Approximate evaluation of marginal association probabilities with belief propagation}

\author{Jason~L.~Williams,~\IEEEmembership{Member,~IEEE} and Roslyn~A.~Lau,~\IEEEmembership{Student, IEEE}
\thanks{Manuscript received September 12, 2012; revised August 12, 2013; released for publication March 28, 2014. Refereeing of this contribution was handled by T.~Luginbuhl.}%
\thanks{The authors are with the National Security, Intelligence, Surveillance and Reconnaissance Division, Defence Science and Technology Organisation, Australia (e-mail: jason.williams@dsto.defence.gov.au, roslyn.lau@dsto.defence.gov.au). Jason Williams is also with the School of Electrical and Electronic Engineering, University of Adelaide, Australia. Roslyn Lau is also with the Research School of Computer Science, Australian National University, Australia.} %
\thanks{Digital Object Identifier 10.1109/TAES.2014.120568}}

\IEEEpubid{\copyright~2014 Crown}

\begin{document}

\maketitle

\begin{abstract}
Data association, the problem of reasoning over correspondence between targets and measurements, is a fundamental problem in tracking. This paper presents a graphical model formulation of data association and applies an approximate inference method, belief propagation (BP), to obtain estimates of marginal association probabilities. We prove that BP is guaranteed to converge, and bound the number of iterations necessary. Experiments reveal a favourable comparison to prior methods in terms of accuracy and computational complexity.
\end{abstract}


%%%%%%%%%%%%%%%%%%%%%%%%%%%%%%%%%%%%%%%%%%%%%%%%%%%%%%%%%%%%%%%%%%%%%%%%%%%%%%%
\section{Introduction}
\label{sec:Introduction}
{\noindent}In recent years, graphical models have emerged as a powerful tool for inference and learning in large scale systems. The promise of graphical models in tracking problems was demonstrated in \cite{CheWai03,CheCet05,CheWai06}. The formulation in the former focused on sensor networks, in which each sensor had a narrow field of view. Non-overlapping regions were defined and association variables were instantiated to hypothesise joint association events for all targets and measurements within a region. Graphical models were studied in a similar application in \cite{GniMih09}.

In the present study, we consider the classical data association problem, in which a single sensor surveils a large number of targets. Each target may give rise to at most one measurement, and each measurement is related to at most one target. We focus on an approximate solution of a core problem in data association: calculating marginal measurement-to-target association probabilities such as those used in joint probabilistic data association (JPDA) \cite{ForBar83}, multi-target mixture reduction \cite{Pao94,RisAru00}, and related methods \cite{HorMas09,Wil12}. 

Calculation of marginal association probabilities is closely related to the computation of the permanent of a non-negative matrix \cite{ColUhl92}, a key problem in the definition of the \emph{\#P-complete} complexity class \cite{Val79}. Brute force, exact calculation of these quantities is intractable for all but the smallest problems. Until recently, practical applications of these methods have relied on simple heuristics; a recent survey of these can be found in \cite{RomCro10}. For example, cheap JPDA \cite{Fit90} replaces the joint weights with a heuristic that maintains the same form of behaviour in some sense (\eg, de-weighting measurements that match several targets). Linear joint integrated PDA (LJIPDA) \cite{MusEva02} provides a heuristic modification of the standard PDA weights to model the impact of other tracks, incorporating an estimate of the probability that each measurement is not used by another track. Linear multitarget IPDA (LMIPDA) \cite{MusLaS08} improves upon this approach, treating measurements from other targets as additional clutter. The approach in \cite{BakAla96} takes an expression which is exact in the two-target case and heuristically extends it to the multi-target case. Other methods are based on pruning of events in the joint weight calculation. Near-optimal JPDA \cite{Roe94} averages over highly likely joint events, first finding the maximum \apost (MAP) event (which can be computed efficiently), then repeating the MAP calculation on modified problems, removing elements chosen in the MAP. JPDA* \cite{BloBlo00} calculates the MAP association conditioned on each subset of targets hypothesised to be detected, and each subset of measurements hypothesised to be target-originated.

\IEEEpubidadjcol

Significant improvements have been made recently towards highly accurate and exact approaches. The efficient hypothesis management (EHM) method \cite{MasBri04,HorMas06} exploits redundancy present in many problems to provide exact evaluation with reduced complexity, effectively providing an improved version of the junction tree algorithm (discussed in \secref{ss:JTree}). However, in some cases (\eg, at the extreme end, dense problems in which each measurement falls within the association gate of each target), complexity remains exponential. Markov chain Monte Carlo data association (MCMCDA) \cite{OhRus09} provides a randomised, fully polynomial time approximation scheme (FPTAS) for estimating the probabilities, but the computational complexity of the method limits its practical use. Subsequent to our initial publications \cite{WilLau10,WilLau10a}, a related correlation decay method \cite{Oh11} has been proposed, providing a deterministic FPTAS. As we show in \secref{sec:Experiments}, the computational complexity of this approach remains problematic.

This paper develops a practical approximation approach to data association based on belief propagation (BP), demonstrating remarkable accuracy, and proving convergence of the algorithm, despite the presence of cycles in the graphical model formulation.

\subsection{Contributions}
\label{ss:Contributions}
{\noindent}In this paper, we examine an emerging method for approximating the marginal association probabilities. Motivated by the success of \cite{BaySha08}, the method was first proposed in \cite{HuaJeb09}, and subsequently studied in \cite{CheKro08,CheKro10}. It was developed independently and evaluated in the tracking context in our preliminary paper \cite{WilLau10}. The contributions of this paper are as follows:
\begin{itemize}
\item Proof of convergence of the method for the most common case in tracking where the probability of detection is non-unity, and the false alarm rate is non-zero. A preliminary version of this result was published in \cite{WilLau10a}. A more general proof (effectively admitting cases with unity probability of detection and/or zero false alarm rate), developed in parallel, was announced in \cite{Von10}, and is available in \cite{Von13}.
\item Analysis of the computational complexity of the method, providing guarantees on the number of iterations required as a function of the problem parameters, and interpretation of these parameters in the tracking context.
\item A thorough experimental evaluation of the accuracy of the approximation in challenging tracking problems, and comparison (including computation time) to other state-of-the-art methods in the tracking literature. The comparison reveals the unique position of the proposed approximation in the accuracy versus computation time trade-off.
\end{itemize}

\subsection{Outline of paper}
\label{ss:Outline}
In \secref{sec:Background}, we introduce the problem of data association and the particular formulation we utilise (\secref{ss:BackDataAssoc}), before introducing the formalism of graphical models (\secref{ss:BackGraphicalModels}). In \secref{sec:BPDA}, we derive our approach, prove convergence, bound computation time, and examine practical stopping criteria. In \secref{sec:Relationships}, we compare and contrast our method to three closely related approaches. In \secref{sec:Experiments}, we present a thorough experimental comparison of our approach to state-of-the-art alternatives.

%%%%%%%%%%%%%%%%%%%%%%%%%%%%%%%%%%%%%%%%%%%%%%%%%%%%%%%%%%%%%%%%%%%%%%%%%%%%%%%
\section{Background}
\label{sec:Background}

\subsection{Data association model}
\label{ss:BackDataAssoc}
{\noindent}We analyse a variant of the classical model (\eg, \cite{ForBar83,BarWil11}) which incorporates uncertainty in target existence using the random finite set (RFS) formalism of \cite{Mah07}. The focus of this paper is approximate calculation of the marginal association probabilities; we describe the model for concreteness, and in order to provide intuition for how the model parameters (\secref{ss:BPComplexInterpret}) impact the number of iterations required for convergence. Details of the model and derivation of RFS filters which use the present algorithm may be found in \cite{Wil12}. We use the abbreviated notation $f(x)$, $f(z|x)$, \etc., to represent the probability density function (PDF) of the continuous random variable corresponding to the value $x$, $z$ conditioned on $x$, \etc., and $p(a)$, $p(a|z)$, \etc., to represent the point mass function (PMF) of the discrete random variable corresponding to the value $a$, $a$ conditioned on $z$, \etc..

Assume that at time $t$ there are $n_t$ targets with states $X_t=\{x_t^1,\dots,x_t^{n_t}\}$ and $m_t$ measurements $Z_t=\{z_t^1,\dots,z_t^{m_t}\}$. We use the symbol $i\in\{1,\dots,n_t\}$ to refer to a target index, and $j\in\{1,\dots,m_t\}$ to refer to a measurement index. Each target may give rise to at most one measurement (with detection probability $\Pd(x_t^i)$), and each measurement may result from at most one target (false alarms occur according to a Poisson point process with intensity $\lfa(z_t^j)$). New targets arrive at each time according to a Poisson point process with intensity $\lambdan(x_t)$.\footnote{For simplicity, we assume that new targets are guaranteed to be detected; the method in \cite{Wil12} shows how this assumption can be relaxed.} Target dynamics models (including death) are not important to the analysis we wish to perform. The measurement likelihood is $f(z_t|x_t)$. The complete set of measurements up to and including time $t$ is denoted as $Z^t = (Z_1,\dots,Z_t)$. Under these assumptions, measurements from either false alarms or new targets will be a Poisson point process with intensity
\[
\lfan(z_t) = \lfa(z_t) + \int{f(z_t|x_t)\lambdan(x_t)\dif x_t}
\]

The relationship between targets and measurements is described via a set of latent association variables, comprising:
\begin{enumerate}
\item For each target $i \in \{1,\dots,n_t\}$, an association variable $a_t^i \in \{0,1,\dots,m_t\}$, the value of which is an index to the measurement with which the target is hypothesised to be associated (zero if the target is hypothesised to have not been detected)
\item For each measurement $j \in \{1,\dots,m_t\}$, an association variable $b_t^j \in \{0,1,\dots,n_t\}$, the value of which is an index to the target with which the measurement is hypothesised to be associated (zero if the measurement is hypothesised to be either a false alarm or a new target)
\end{enumerate}
Note that the two sets of association variables are entirely redundant: given the information from either set, the other can be reconstructed perfectly. As we will see in the following sections, this choice of formulation results in an approximate algorithm with guaranteed convergence, and remarkable accuracy.

For clarity, we neglect uncertainty of target existence in this introduction; as shown in \cite{Wil12}, incorporation of this phenomenon results in the slightly modified weights in \eqref{eq:WeightInput}. Assuming that the prior distribution (conditioned on previous measurements) factors, \ie, 
\begin{equation}\label{eq:PriorFactoring}
f(x_t^1,\dots,x_t^{n_t}|Z^{t-1})\approx\prod_{i=1}^{n_t}f(x_t^i|Z^{t-1})
\end{equation}
the joint distribution of target states at time $t$, association variables at time $t$ and measurements at time $t$ is:
\ifCLASSOPTIONdraftcls
\begin{multline}
f(x_t^1,\dots,x_t^{n_t},a_t^1,\dots,a_t^{n_t},b_t^1,\dots,b_t^{m_t},Z_t|Z^{t-1}) = %\\
\gamma(a,b)\cdot\left[\prod_{i|a_t^i>0} \Pd(x_t^i)f(z_t^{a_t^i}|x_t^i)f(x_t^i|Z^{t-1})\right]\cdot\\
\cdot\left[\prod_{i|a_t^i=0} [1-\Pd(x_t^i)]f(x_t^i|Z^{t-1})\right] %\cdot\\
\cdot \left[\exp\left\{{\textstyle-\int{\lfan(z)\dif z}}\right\}\cdot\prod_{j|b_t^j=0} \lfan(z_t^j)\right]
\end{multline}
\else
\begin{multline}
f(x_t^1,\dots,x_t^{n_t},a_t^1,\dots,a_t^{n_t},b_t^1,\dots,b_t^{m_t},Z_t|Z^{t-1}) = \\
\gamma(a,b)\cdot\left[\prod_{i|a_t^i>0} \Pd(x_t^i)f(z_t^{a_t^i}|x_t^i)f(x_t^i|Z^{t-1})\right]\cdot\\
\cdot\left[\prod_{i|a_t^i=0} [1-\Pd(x_t^i)]f(x_t^i|Z^{t-1})\right]\cdot\\
\cdot \left[\exp\left\{{\textstyle-\int{\lfan(z)\dif z}}\right\}\cdot\prod_{j|b_t^j=0} \lfan(z_t^j)\right]
\end{multline}
\fi
where $\gamma(a,b)=1$ if $(a_t^i)$ and $(b_t^j)$ form a consistent association event, (\ie, if $a_t^i=j>0$ then $b_t^j=i$ and vice versa) and $\gamma(a,b)=0$ otherwise. The posterior distribution of target state and association is then
\ifCLASSOPTIONdraftcls
\begin{multline}
f(x_t^1,\dots,x_t^{n_t},a_t^1,\dots,a_t^{n_t},b_t^1,\dots,b_t^{m_t}|Z^t) \propto %\\
\gamma(a,b)\cdot\left[\prod_{i|a_t^i>0} \Pd(x_t^i)f(z_t^{a_t^i}|x_t^i)f(x_t^i|Z^{t-1})\right]\cdot\\
\cdot\left[\prod_{i|a_t^i=0} [1-\Pd(x_t^i)]f(x_t^i|Z^{t-1})\right]
\cdot \left[\prod_{j|b_t^j=0} \lfan(z_t^j)\right]
\end{multline}
\else
\begin{multline}
f(x_t^1,\dots,x_t^{n_t},a_t^1,\dots,a_t^{n_t},b_t^1,\dots,b_t^{m_t}|Z^t) \propto \\
\gamma(a,b)\cdot\left[\prod_{i|a_t^i>0} \Pd(x_t^i)f(z_t^{a_t^i}|x_t^i)f(x_t^i|Z^{t-1})\right]\cdot\\
\cdot\left[\prod_{i|a_t^i=0} [1-\Pd(x_t^i)]f(x_t^i|Z^{t-1})\right]
\cdot \left[\prod_{j|b_t^j=0} \lfan(z_t^j)\right]
\end{multline}
\fi
Marginalising out target states and dividing by $\prod_i[1-\Pd(x_t^i)]f(x_t^i|Z^{t-1})\prod_j\lfan(z_t^j)$, we find the posterior distribution of association to be
\ifCLASSOPTIONdraftcls
\begin{equation}
f(a_t^1,\dots,a_t^{n_t},b_t^1,\dots,b_t^{m_t}|Z^t) \propto 
\gamma(a,b)\cdot\prod_{i|a_t^i>0} \frac{
\int{\Pd(x_t^i)f(z_t^{a_t^i}|x_t^i)f(x_t^i|Z^{t-1})\dif x_t^i}
}{
\int{[1-\Pd(x_t^i)]f(x_t^i|Z^{t-1})\dif x_t^i}\lfan(z_t^{a_t^i})
}
\end{equation}
\else
\begin{multline}
f(a_t^1,\dots,a_t^{n_t},b_t^1,\dots,b_t^{m_t}|Z^t) \propto \\
\gamma(a,b)\cdot\prod_{i|a_t^i>0} \frac{
\int{\Pd(x_t^i)f(z_t^{a_t^i}|x_t^i)f(x_t^i|Z^{t-1})\dif x_t^i}
}{
\int{[1-\Pd(x_t^i)]f(x_t^i|Z^{t-1})\dif x_t^i}\lfan(z_t^{a_t^i})
}
\end{multline}
\fi
JPDA and related methods seek to approximate the posterior distribution
\ifCLASSOPTIONdraftcls
\begin{equation}
f(x_t^1,\dots,x_t^{n_t}|Z^t) = 
\sum_{a_t^1,\dots,a_t^{n_t}}p(a_t^1,\dots,a_t^{n_t}|Z^t)f(x_t^1,\dots,x_t^{n_t}|a_t^1,\dots,a_t^{n_t},Z^t)
\end{equation}
\else
\begin{multline}
f(x_t^1,\dots,x_t^{n_t}|Z^t) = \\ 
\sum_{a_t^1,\dots,a_t^{n_t}}p(a_t^1,\dots,a_t^{n_t}|Z^t)f(x_t^1,\dots,x_t^{n_t}|a_t^1,\dots,a_t^{n_t},Z^t)
\end{multline}
\fi
If, as assumed, the prior factors, then 
\begin{equation}
f(x_t^1,\dots,x_t^{n_t}|a_t^1,\dots,a_t^{n_t},Z^t) = \prod_{i=1}^{n_t}f(x_t^i|a_t^i,Z^t)
\end{equation}
Suppose the joint distribution of association events is approximated by the product of its marginals; as shown in \cite[p277]{KolFri09}, this is the product form distribution which minimises the Kullback-Leibler divergence:
\begin{equation}\label{eq:AssocFactorApprox}
p(a_t^1,\dots,a_t^{n_t}|Z^t) \approx \prod_{i=1}^{n_t}p(a_t^i|Z^t)
\end{equation}
Subsequently, with no further approximation, the full posterior is:
\begin{equation}
f(x_t^1,\dots,x_t^{n_t}|Z^t) \approx \prod_{i=1}^{n_t}\sum_{a_t^i}p(a_t^i|Z^t)f(x_t^i|a_t^i,Z^t)
\end{equation}
This standard approach, pioneered in \cite{ForBar83}, can be extended to incorporate target existence similarly as in \cite{Wil12}, or using the closely related methods \cite{MusEva04}. The hypothesis-conditioned updated distribution $f(x_t^i|a_t^i,Z^t)$ may be calculated using standard state estimation methods such as the Kalman filter \cite{Kal60,May79}, unscented Kalman filter \cite{JulUhl04} or particle filter \cite{GorSal93,RisAru04} depending on the model in use. The challenging problem is computation of the marginal probabilities $p(a_t^i|Z^t)$; this computation is believed to be \#P complete. Under the assumptions stated, the joint distribution can be expressed as
\ifCLASSOPTIONdraftcls
\begin{equation}
\label{eq:JointDistribution}
p(a_t^1,\dots,a_t^{n_t},b_t^1,\dots,b_t^{m_t}|Z^t) \propto
\prod_{i=1}^{n_t}\left\{
\psi_i(a_t^i)
\prod_{j=1}^{m_t}\psi_{i,j}(a_t^i,b_t^j)
\right\}
\end{equation}
\else
\begin{multline}
\label{eq:JointDistribution}
p(a_t^1,\dots,a_t^{n_t},b_t^1,\dots,b_t^{m_t}|Z^t) \propto \\ 
\prod_{i=1}^{n_t}\left\{
\psi_i(a_t^i)
\prod_{j=1}^{m_t}\psi_{i,j}(a_t^i,b_t^j)
\right\}
\end{multline}
\fi
where, collectively the functions
\begin{equation}
\psi_{i,j}(a_t^i,b_t^j) = \begin{cases}
0, & a_t^i = j, b_t^j \neq i \mbox{ or } b_t^j = i, a_t^i \neq j \\
1, & \mbox{otherwise}
\end{cases}
\end{equation}
provide a factored form of $\gamma(a,b)$, collectively ensuring that the redundant sets of association variables $(a_t^1,\dots,a_t^n)$ and $(b_t^1,\dots,b_t^{m_t})$ are consistent, \ie, that any event in which the collections are inconsistent has zero probability. An example of an inconsistent event is one in which the target association variable $a_t^i$ indicates that target $i$ is associated with measurement $j$, but the measurement association variable $b_t^j$ does not indicate that measurement $j$ is associated with target $i$; this implicitly excludes cases in which the same measurement is associated with two targets, or the same target is associated with two measurements. The factors $\psi_i$ encode the problem data, with $\psi_i(a_t^i=0)=1$ and for $a_t^i=j>0$,
\begin{equation}
\psi_i(j) = 
\frac{
r^i_t\int{\Pd(x_t^i)f(z_t^j|x_t^i)f(x_t^i|Z^{t-1})\dif x_t^i}
}{
\lfan(z_t^j)\left[1 - r^i_t\int{\Pd(x_t^i)f(x_t^i|Z^{t-1})\dif x_t^i}\right]
} \label{eq:WeightInput}
\end{equation}
where $r^i_t$ is the probability of existence of the $i$-th track prior to incorporation of the new measurement.\footnote{We do not consider update equations for this quantity, hence we do not require notation for the post-update probability of existence.} We reinforce that the details of this model are not the topic of this paper; for the present context, the model is a minor variation of previous, standard works such as \cite{ForBar83,RisAru00,HorMas09}. The same problem of estimating marginal association probabilities arises in other formulations of tracking problems such as extensions of the probabilistic multiple hypothesis tracker (PMHT) \cite{RagWil95,LauWil13}. 

The topic of interest in the current paper is how we might tractably estimate the marginal association probabilities:
\begin{align}
p(a_t^i|Z^t) &= \sum_{a_t^{i'},\, i'\neq i; \;b_t^j\,\forall\,j}p(a_t^1,\dots,a_t^{n_t},b_t^1,\dots,b_t^{m_t}|Z^t) \label{eq:MarginalA} \\
p(b_t^j|Z^t) &= \sum_{a_t^i\,\forall\,i;\; b_t^{j'},\, j'\neq j}p(a_t^1,\dots,a_t^{n_t},b_t^1,\dots,b_t^{m_t}|Z^t) \label{eq:MarginalB}
\end{align}
Note that in the standard problem setup (\eg, JPDA), the approximation of \eqref{eq:AssocFactorApprox} has already been applied recursively at each prior time step. We focus on the calculation of \eqref{eq:MarginalA} and \eqref{eq:MarginalB} in the current time step, where the joint is given by \eqref{eq:JointDistribution}. As such, we refer to methods which correctly evaluate these as being exact, and methods which estimate these quantities with some error as being approximate.

\subsection{Graphical models}
\label{ss:BackGraphicalModels}
{\noindent}Graphical models \cite{Lau96,WaiJor08,KolFri09} aim to represent and manipulate the joint probability distributions of many variables efficiently by exploiting factorisation. The Kalman filter \cite{Kal60} and the hidden Markov model (HMM) \cite{Rab89} are two examples of algorithms that exploit sparsity of a particular kind (\ie, a Markov chain) to efficiently conduct inference on systems involving many random variables. Inference methods based on the graphical model framework generalise these algorithms to a wider variety of state spaces and dependency structures. 

Graphical model methods have been developed for undirected graphical models (Markov random fields), directed graphical models (Bayes nets) and factor graphs. In this work we consider a subclass of pairwise undirected models, involving nodes (\ie, random variables) $n\in\calN$, and edges (\ie, dependencies) $e\in\calE\subset\calN\times\calN$, and where the joint distribution can be written as:\footnote{In the general setting, the joint distribution is a product of maximal cliques \cite[p9]{WaiJor08}. Since the graph is undirected, we assume that $\calE$ is symmetric, \ie, if $(i,j)\in\calE$ then $(j,i)\in\calE$. We need only incorporate one of these two factors in the distribution.}
\[
p(x_{\calN}) \propto \prod_{n\in\calN}\psi_n(x_n)\prod_{(i,j)\in\calE}\psi_{i,j}(x_i,x_j)
\]
It should be immediately clear that \eqref{eq:JointDistribution} is in this form. As another example, a Markov chain involving variables $(x_1,\dots,x_T)$ may be formulated by setting $\psi_1(x_1) = p(x_1)$ for the initial prior, $\psi_t(x_t)=1$ for $t>1$, and edges $\psi_{t-1,t}(x_{t-1},x_t) = p(x_t|x_{t-1})$, $t\in\{2,\dots,T\}$ representing the Markov transition kernels, although other formulations are possible. 

Optimal inference can be conducted on tree-structured graphs using belief propagation (BP). BP proceeds by passing messages between neighbouring nodes. We denote by $\mu_{i\rightarrow j}(x_j)$ the message sent from node $i\in \calN$ to node $j\in\calN$ where $(i,j)\in\calE$. The iterative update equations are then:
\begin{equation}
\mu_{i\rightarrow j}(x_j) \propto \sum_{x_i}\psi_{i,j}(x_i,x_j)\psi_i(x_i) \prod_{(j',i)\in\calE, j'\neq j}\mu_{j'\rightarrow i}(x_i)
\label{eq:BPMessage}
\end{equation}
For obvious reasons, this is also known as the \emph{sum-product} algorithm. If the summations are replaced with maximisation operations, then we arrive at max-product BP, which generalises the well-known Viterbi algorithm \cite{Vit67}, providing the MAP joint state of all variables in the graph. At convergence of sum-product BP, the marginal distribution at a node $n$ can be calculated as:
\begin{equation}
p(x_n) \propto \psi_n(x_n)\prod_{(n,i)\in\calE} \mu_{i\rightarrow n}(x_n)\label{eq:BPMarginal}
\end{equation}
In the case of a Markov chain, if all nodes are jointly Gaussian, then BP is equivalent to a Kalman smoother. Similarly, if all nodes are discrete, then BP is equivalent to inference on an HMM using the forward-backward algorithm. BP unifies these algorithms, and extends them from chains to trees. 

Inference in cyclic graphs (graphs that have cycles, \ie, that are not tree-structured) is far more challenging. Conceptually, one can always convert an arbitrary cyclic graph to a tree by merging nodes (\eg, so-called \emph{junction tree} representations) \cite{Lau96,KolFri09}, but in practical problems, the dimensionality of the agglomerated variables may be prohibitive. BP may be applied to cyclic graphs; practically, this simply involves repeated application of \eqref{eq:BPMessage} until convergence occurs (\ie, until the maximum change between subsequent messages is less than a pre-set threshold). Unfortunately, this is neither guaranteed to converge to the right answer, nor to converge at all. Nevertheless, and perhaps surprisingly, it has exhibited excellent empirical performance in many practical problems \cite{MurWei99}. For example, the popular iterative turbo decoding algorithm has been shown to be an instance of BP applied to a cyclic graph \cite{McEMac98}. 

The current understanding of BP in cyclic graphs stems largely from \cite{YedFre05}. It has been shown (\eg, \cite[Theorem 3.4]{WaiJor08}) that one can recover exact marginal probabilities from an optimisation of a convex function known as the Gibbs free energy, one term of which is the joint entropy of the distribution. While this optimisation is intractable, it points to a family of \emph{variational inference} methods that approximate the objective function and the feasible set to enable calculation of marginal probability estimates without ever manipulating the full joint distribution \cite{Jaa00,WaiJor08}. It has been shown \cite{YedFre05} that BP (when it converges) solves one such approximation, in which the objective is approximated by the Bethe free energy (replacing the joint entropy with a series of differences of pairwise edge entropies and node entropies), and the feasible set (\ie, the set of all valid probability distributions) is approximated as the distributions for which the pairwise joint distributions along edges are consistent with node marginals \cite[4.1.1]{WaiJor08}. The BP message iterates can be viewed as a general iterative method for solving a series of fixed point equations derived from the optimality conditions of the Bethe free energy variational problem \cite[4.1.3]{WaiJor08}. The marginal probability estimates obtained using BP are referred to as \emph{beliefs}.

The two approximations made by BP each lead to difficulties in certain circumstances. Firstly, whereas the Gibbs free energy is convex (since entropy is concave), the Bethe free energy is, in general, neither concave nor convex. The practical behaviour is that errors are often either very small or very large, and the general intuition is that failure of convexity leads to large errors; this has led to new families of algorithms such as tree re-weighted sum product \cite{WaiJaa05}, which replace the Bethe free energy with a convex surrogate. Secondly, the approximation of the feasible set may admit points which do not correspond to any valid probability distribution, hence the solution may be infeasible. Methods seeking to tighten the feasible set include \cite{WaiJor06}.

In the case in which the graph is a tree, the Bethe free energy is equivalent to the Gibbs free energy (once the domains have been suitably identified) and hence convex, and the feasible set is exact. Not surprisingly, then, BP is exact on trees. In recent years, a number of other cases have emerged in which there are guarantees on aspects of the algorithm; this paper studies one of these cases.

\subsection{Graphical models, matching and permanents}
\label{ss:BackMatching}
Matching problems involve graphs in which pairs of nodes (connected by edges) are matched or unmatched. A subset of edges is called a matching (\ie, an allowable configuration) if no two edges from the matching are incident on the same node. A maximal matching is one in which there is no additional edge that can be matched while maintaining a matching. Finding the most likely data association corresponds to an assignment problem, which can be formulated as a maximum weighted matching problem on a bipartite graph (\ie, each edge has a weight, \ie, the log of the association weight \eqref{eq:WeightInput}, and we seek the set of edges which maximises the sum of the weights).

In recent years, several authors have applied graphical models to matching problems. In \cite{BaySha08}, it was shown that max-product BP can be used to optimally solve assignment problems (\ie, maximum weighted bipartite matching) in time comparable to the well-known auction algorithm. In \cite{SanMal11}, this result was extended, showing that max-product converges (finding the optimal solution) in general matching problems if and only if the LP relaxation has a unique integral optimum.

In this paper, we study calculation of marginal probabilities on the same model as \cite{BaySha08}, and utilise the same graph formulation. The problem of counting the number of matchings is a special case of evaluation of marginal probabilities, in which all joint events with non-zero probability (\ie, all matchings) are equally likely. The cavity method for counting matchings is studied in \cite{BayNai06}, and convergence of the cavity method is proven on general graphs. The proof of convergence of the current work is closely related to this method; in essence \secref{ss:BPConvergence} treats the weighted version of the problem considered in \cite{BayNai06} on a subclass of graphs (specifically, bipartite graphs in which for each node there is a non-zero weight that no neighbouring edge is matched).

 In \cite{Von13}, it was shown that, for a similar model involving perfect matchings, the Bethe free energy can be parameterised by the doubly stochastic matrix of marginal target-measurement probabilities, and is convex with respect to this parameterisation. Furthermore, the Birkhoff-von Neumann theorem states that any doubly-stochastic matrix is a convex combination of permutations \cite{Von13} (\eg, different association configurations), hence infeasibility is also ruled out.\footnote{\textit{I.e.}, there is guaranteed to be a distribution over joint associations that yields the beliefs that BP obtains.} In the following section, we state the model, and derive results relating to convergence and complexity of BP in this model.

%%%%%%%%%%%%%%%%%%%%%%%%%%%%%%%%%%%%%%%%%%%%%%%%%%%%%%%%%%%%%%%%%%%%%%%%%%%%%%%
\section{Belief propagation data association}
\label{sec:BPDA}
{\noindent}In this section, we consider the use of BP to approximate the association probabilities in \eqref{eq:JointDistribution}. We commence in \secref{ss:BPFormulation} by explicitly stating the BP update equations, and then (in \secref{ss:BPSimplified}) obtain an equivalent form that reduces computation complexity from $O(n_t^3 m_t^2 + n_t^2 m_t^3)$ per iteration to $O(n_t m_t)$. The reduction is effectively a sum-product version of the simplified algorithm in \cite{BaySha08}, and an equivalent form was provided in \cite{HuaJeb09}. In \secref{ss:BPConvergence} we use the simplified update equations to prove convergence of the algorithm, and then, in \secref{ss:BPComplexity}, bound the number of iterations required for convergence. In \secref{ss:BPComplexInterpret} we calculate the expected value of the quantity involved in the bound on the number of iterations required, providing intuition into the practical behaviour of the method. In \secref{ss:BPStoppingCrit} we provide a criterion for terminating the BP computation with a guarantee of the deviation from the converged solution.

\subsection{Formulation}
\label{ss:BPFormulation}
{\noindent}The model we study is that of \eqref{eq:JointDistribution}. As illustrated in \figref{fig:Bipartite}, this is a bipartite model in which all target association variables $(a_t^i)_{i\in\{1,\dots,n_t\}}$ are connected to all measurement association variables $(b_t^j)_{j\in\{1,\dots,m_t\}}$. In this case, BP may be implemented via two half-iterations, alternating between the two sets of messages, $(\mu_{a_t^i\rightarrow b_t^j})$ and $(\mu_{b_t^j\rightarrow a_t^i})$; abbreviating notation, we refer to these as $(\mu_{i\rightarrow j})$ and $(\nu_{j\rightarrow i})$ respectively. Omitting the time index from $a_t^i$ and $b_t^j$ for simplicity, the message update equations are:
\begin{figure}[t]
\centering
\includegraphics{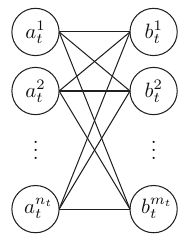}
\caption{Graphical model formulation employed for data association. The value of the random variable $a_t^i$ is the index of the measurement with which target $i$ is hypothesised to be associated, while the value of the random variable $b_t^j$ is the index of the target with which measurement $j$ is hypothesised to be associated.}
\label{fig:Bipartite}
\end{figure}
\begin{align}
\mu_{i\rightarrow j}(b^j) &= \sum_{a^i} \psi_i(a^i)\psi_{i,j}(a^i,b^j) \prod_{j'\neq j}\nu_{j'\rightarrow i}(a^i) \label{eq:BPTgtToMeas}\\
&= \begin{cases}
\psi_i(j)\prod_{j'\neq j}\nu_{j'\rightarrow i}(j), & b^j = i \\
\sum_{a^i\neq j}\psi_i(a^i)\prod_{j'\neq j}\nu_{j'\rightarrow i}(a^i), & b^j \neq i
\end{cases} \label{eq:BPTgtToMeas2} \\
\nu_{j\rightarrow i}(a^i) &= \sum_{b^j} \psi_{i,j}(a^i,b^j) \prod_{i'\neq i}\mu_{i'\rightarrow j}(b^j)\label{eq:BPMeasToTgt} \\
&= \begin{cases}
\prod_{i'\neq i}\mu_{i'\rightarrow j}(i), & a^i = j \\
\sum_{b^j\neq i}\prod_{i'\neq i}\mu_{i'\rightarrow j}(b^j), & a^i \neq j
\end{cases} \label{eq:BPMeasToTgt2}
\end{align}
Na\"{i}vely, the complexity per iteration of this procedure is $O(n_t^3 m_t^2 + n_t^2 m_t^3)$, since each iteration involves sending a message from each of the $n_t$ target association variables to each of the $m_t$ measurement association variables (and vice versa), and each message involves $(m_t+1)$ (respectively $(n_t+1)$) values, each of which requires $O(n_t^2)$ (respectively $O(m_t^2)$) calculations.

\subsection{Simplified algorithm}
\label{ss:BPSimplified}
{\noindent}As shown in \cite{BaySha08,HuaJeb09,WilLau10a}, these computations can be dramatically simplified by observing that in \eqref{eq:BPTgtToMeas2} (and \eqref{eq:BPMeasToTgt2}), while the message consists of $(n_t+1)$ (respectively $(m_t+1)$) values there are only two distinct values (\ie, $b^j=i$ and $b^j\neq i$ in \eqref{eq:BPTgtToMeas2}, and $a^i=j$ and $a^i\neq j$ in \eqref{eq:BPMeasToTgt2}). Further, since we are free to renormalise messages, we may divide by one of these two values ($\mu_{i\rightarrow j}(b^j\neq i)$ and $\nu_{j\rightarrow i}(a^i\neq j)$ respectively) to obtain a scalar representation of the message. Using the shorthand $\sum_{j'\neq j, j'>0}$ for the sum over the set $\{1,\dots,j-1,j+1,\dots,m_t\}$, and similarly $\sum_{i'\neq i, i'>0}$ for the sum over the set $\{1,\dots,i-1,i+1,\dots,n_t\}$, the message update equations in terms of these scalars become:
\begin{align}
\mu_{i\rightarrow j} &= \frac{\psi_i(j)}{1 + \sum_{j'\neq j, j'>0}\psi_i(j')\nu_{j'\rightarrow i}}\label{eq:FixedPoint1}\\
\nu_{j\rightarrow i} &= \frac{1}{1 + \sum_{i'\neq i, i'>0}\mu_{i'\rightarrow j}}\label{eq:FixedPoint2}
\end{align}
where we have exploited the choice that $\psi_i(a^i=0)=1$ to reinforce that the denominator is greater than zero. Upon convergence, the approximate marginal association probabilities (beliefs) can be obtained by 
\begin{align}
\hat{p}(a^i=j|Z^t) & = \frac{\psi_i(j)\nu_{j\rightarrow i}}{\sum_{j'}\psi_i(j')\nu_{j'\rightarrow i}} \label{eq:Marginal1}\\
\hat{p}(b^j=i|Z^t) & = \frac{\mu_{i\rightarrow j}}{\sum_{i'}\mu_{i'\rightarrow j}} \label{eq:Marginal2}
\end{align}
where the shorthand $\sum_{i'}$ is the sum $\sum_{i'=0}^{n_t}$, similarly $\sum_{j'}$ is the sum $\sum_{j'=0}^{m_t}$, $\nu_{0\rightarrow i}\triangleq 1$ and $\mu_{0\rightarrow j}\triangleq 1$. An $O(n_t m_t)$ (per iteration) Matlab implementation of these update equations was given in \cite{WilLau10a}.

\subsection{Proof of convergence}
\label{ss:BPConvergence}
{\noindent}Most previous work on the convergence of BP (such as \cite{IhlFis05a}) requires the factors of the graph to satisfy conditions on dynamic range (\eg, $\max_{a^i,b^j}\psi_{i,j}(a^i,b^j)/\min_{a^i,b^j}\psi_{i,j}(a^i,b^j)$). These methods do not apply to the present case since the factor $\psi_{i,j}(a^i,b^j)$ has infinite dynamic range. Nevertheless, we show here that it is possible to establish that the simplified expressions in \eqsref{eq:FixedPoint1} and \noeqref{eq:FixedPoint2} are contractions, thus guaranteeing convergence of the method. The property we exploit in the proof is the same as the correlation decay property utilised in the recent work \cite{Oh11}.

Let $\muv = \gv(\nuv)$ be the update defined in \eqref{eq:FixedPoint1} in vector form with $\nuv=(\nu_{j\rightarrow i})_{i\in\{1,\dots,n_t\},j\in\{1,\dots,m_t\}}$, $\muv=(\mu_{i\rightarrow j})_{i\in\{1,\dots,n_t\},j\in\{1,\dots,m_t\}}$, and $\nuv = \hv(\muv)$ be the update defined in \eqref{eq:FixedPoint2}. The domain of $\gv(\cdot)$ and $\hv(\cdot)$ is considered to be $[0,\infty)^{n_t m_t}$, the range of $\hv(\cdot)$ is $(0,1]^{n_t m_t}$, and the range of $\gv(\cdot)$ is the Cartesian product $\prod_{i,j}\calG_{ij}$, where
\begin{equation}
\calG_{ij} = \begin{cases}
\{0\}, & \psi_i(j) = 0 \\
(0,\psi_i(j)], & \mbox{otherwise}
\end{cases}
\end{equation}
Note that an element $\mu_{i\rightarrow j}$ resulting from $\gv(\cdot)$ is zero if and only if the corresponding value $\psi_i(j)$ is zero, in which case the range of that element $\calG_{ij}$ is the single point $\{0\}$, and all subsequent iterates are also zero.

A function $f : \calX \rightarrow \calX$ operating in a metric space $\calX$ with distance metric $d : \calX \times \calX \rightarrow \Rbb_{\geq 0}$ is a \emph{contraction} if there is a contraction factor $\alpha<1$ such that $d(f(x),f(y))\leq \alpha d(x,y)$ $\forall$ $x,y\in \calX$ \cite[9.22]{Rud76}. If $f$ is a contraction on a complete metric space, then any sequence resulting from repeated application of $f$ will converge to the same fixed point \cite[9.23]{Rud76}. Following \cite{IhlFis05a}, we use the distance metric
\begin{equation}\label{eq:DistanceMetric}
d(\muv,\tilde{\muv}) = \max_{i,j}\left|
\log\frac{\mu_{i\rightarrow j}}{\tilde{\mu}_{i\rightarrow j}}
\right|
\end{equation}
where we use the convention $\frac{0}{0}=1$ (as discussed above, the point zero is excluded from the range of the function except for elements for which the range is the single point $\{0\}$). To show that this satisfies the triangle inequality, note that
\begin{align*}
d(\muv,\tilde{\muv}) &= \max_{i,j}\left|
\log\frac{\mu_{i\rightarrow j}}{\nu_{i\rightarrow j}}\cdot\frac{\nu_{i\rightarrow j}}{\tilde{\mu}_{i\rightarrow j}}
\right| \\
&\leq \max_{i,j}\left|
\log\frac{\mu_{i\rightarrow j}}{\nu_{i\rightarrow j}}\right| + 
\max_{i,j}\left|
\log\frac{\nu_{i\rightarrow j}}{\tilde{\mu}_{i\rightarrow j}} \right| \\
&= d(\muv,\nuv) + d(\nuv,\tilde{\muv})
\end{align*}
Other properties of a distance metric are trivially satisfied.

We now prove a preliminary result relating to the form of the contraction factor that we use.
\begin{lemma}\label{lem:ContractionFactor}
For $L>1$ and $c>0$, the function
\begin{equation}
\alpha(L,c) = \frac{\log\left(\frac{1+cL}{1+c}\right)}{\log L}
\label{eq:ConvProofContraction}
\end{equation}
is strictly less than one, and is monotonically increasing in $L$.
\end{lemma}
\begin{proof}
First note that
\[
\frac{1+cL}{1+c}=L\frac{\frac{1}{L}+c}{1+c} < L
\]
This gives the first result. For the second result, it suffices to show that $\frac{\dif}{\dif L}\log\alpha(L,c) \geq 0$, \ie, that
\[
\left(\frac{1+cL}{1+c}\right)\log\left(\frac{1+cL}{1+c}\right) \leq \frac{c}{1+c} L\log{L}
\]
This result is an immediate consequence of convexity of $x\log{x}$, since $\frac{1+cL}{1+c} = \frac{1}{1+c}1 + \frac{c}{1+c}L$.
\end{proof}

The following two lemmas establish that the BP updates in \eqsref{eq:FixedPoint1} and \noeqref{eq:FixedPoint2} are contractions.

\begin{lemma}\label{lem:BPFixedPoint1Contraction}
For all $(\nuv,\tilde{\nuv})$ with $d(\nuv,\tilde{\nuv})\leq\log\bar{L}$, the message update $\gv(\cdot)$ is a contraction with respect to the distance metric $d(\cdot,\cdot)$ with factor $\alpha(\bar{L},W_*)$, where
\[
W_i = \sum_{a^i>0}\psi_i(a^i), \quad W_* = \max_i W_i
\]
\end{lemma}
\begin{proof}
Assume that, $\forall\; (i,j)$, $0 < \nu_{j\rightarrow i} \leq 1$ and $0<\tilde{\nu}_{j\rightarrow i} \leq 1$; this is not restrictive as it is guaranteed to be satisfied for any $(\nuv,\tilde{\nuv})$ resulting from \eqref{eq:FixedPoint2}. Let
\[
L \triangleq \exp d(\nuv,\tilde{\nuv}) \leq \bar{L} < \infty
\]
Then $\nu_{j\rightarrow i} \leq L\tilde{\nu}_{j\rightarrow i}$ and $\tilde{\nu}_{j\rightarrow i} \leq L\nu_{j\rightarrow i}$. If $\psi_i(j)=0$ then $g_{i,j}(\nuv)=g_{i,j}(\tilde{\nuv})=0$, and $g_{i,j}(\nuv)/g_{i,j}(\tilde{\nuv})\triangleq 1$. Otherwise, consider the quotient 
\begin{align*}
\frac{g_{i,j}(\nuv)}{g_{i,j}(\tilde{\nuv})} &= \frac{1+\sum_{j'\neq j,j'>0}\psi_i(j')\tilde{\nu}_{j'\rightarrow i}}{1+\sum_{j'\neq j,j'>0}\psi_i(j')\nu_{j'\rightarrow i}} \\
&\leq \frac{1+\sum_{j'\neq j,j'>0}\psi_i(j')L\nu_{j'\rightarrow i}}{1+\sum_{j'\neq j,j'>0}\psi_i(j')\nu_{j'\rightarrow i}} \\
&= \frac{1 + c_{ij}L}{1+c_{ij}} \leq \frac{1 + W_* L}{1+ W_*} \\
&= L^{\alpha(L,W_*)} \leq L^{\alpha(\bar{L},W_*)} 
\intertext{where $c_{ij}=\sum_{j'\neq j,j'>0}\psi_i(j')\nu_{j'\rightarrow i}\leq W_i \leq W_*$, and the final step uses the second result of \lemref{lem:ContractionFactor}. Following similar steps,}
\frac{g_{i,j}(\tilde{\nuv})}{g_{i,j}(\nuv)} &\leq \frac{1+W_* L}{1+W_*} \leq L^{\alpha(\bar{L},W_*)}
\end{align*}
Combining the two cases and taking the logarithm,
\begin{equation}
\left|\log\frac{g_{i,j}(\nuv)}{g_{i,j}(\tilde{\nuv})}\right| \leq \alpha(\bar{L},W_*) d(\nuv,\tilde{\nuv}) \; \forall \; i,j
\end{equation}
Since this applies for each $(i,j)$, it also applies to the maximum over $(i,j)$. Thus, by definition of the distance metric in \eqref{eq:DistanceMetric},
\begin{equation}
d(\gv(\nuv),\gv(\tilde{\nuv})) \leq \alpha(\bar{L},W_*) d(\nuv,\tilde{\nuv})
\end{equation}
which is the desired result.
\end{proof}

We now show the same result for the equation in the alternative step, \eqref{eq:FixedPoint2}.
\begin{lemma}\label{lem:BPFixedPoint2Contraction}
For all $(\muv,\tilde{\muv})$ with $d(\muv,\tilde{\muv})\leq\log\bar{L}$, the message update $\hv(\cdot)$ is a contraction with respect to the distance metric $d(\cdot,\cdot)$ with factor $\alpha(\bar{L},W^*)$, where
\[
W^j = \sum_{i>0}\psi_i(j), \quad W^* = \max_j W^j
\]

\end{lemma}
\begin{proof}
Assume that, $\forall\; (i,j)$, $0 \leq \mu_{i\rightarrow j} \leq \psi_i(j)$ and $0\leq\tilde{\mu}_{i\rightarrow j} \leq \psi_i(j)$, and that $\mu_{i\rightarrow j}=0$ or $\tilde{\mu}_{i\rightarrow j}=0$ if and only if $\psi_i(j)=0$ (in which case $\mu_{i\rightarrow j}/\tilde{\mu}_{i\rightarrow j}\triangleq 1$). Again, this is not restrictive as it is guaranteed to be satisfied for any $(\muv,\tilde{\muv})$ resulting from \eqref{eq:FixedPoint1}. Consequently, $\sum_i\mu_{i\rightarrow j}\leq W^j\leq W^*$ (and similarly for $\tilde{\muv}$). Let
\[
L \triangleq \exp d(\muv,\tilde{\muv}) \leq \bar{L} < \infty
\]
Then, following the same steps as in the previous proof,
\begin{align*}
\frac{h_{i,j}(\muv)}{h_{i,j}(\tilde{\muv})} &= \frac{1+\sum_{i'\neq i}\tilde{\mu}_{i'\rightarrow j}}{1+\sum_{i'\neq i}\mu_{i'\rightarrow j}} \\
&\leq \frac{1 + W^* L}{1+W^*} \leq L^{\alpha(\bar{L},W^*)}\\
\frac{h_{i,j}(\tilde{\muv})}{h_{i,j}(\muv)} &\leq \frac{1 + W^* L}{1+W^*} \leq L^{\alpha(\bar{L},W^*)}
\end{align*}
Similar to the previous lemma, this is true for all $(i,j)$, thus we take logarithms, combine cases, and obtain the desired result.
\end{proof}

\begin{theorem}
The loopy BP message update maps $\gv(\cdot)$, $\hv(\cdot)$ converge to the same stationary point regardless of the initialisation.
\end{theorem}

We omit the proof of the theorem since it is a straight-forward application of contraction mapping results; see \cite[Sect 9.23]{Rud76}, and note that the message iterates are contained in a compact subset of the space (\ie, with elements either being zero for all $k$, or being bounded below and above by strictly positive finite constants) hence establishing completeness. 

Note that in most graph structures, it would be necessary to establish that the overall BP message update operation is a contraction. Since the graph in this case is bipartite, messages $\mu_{a_t^i\rightarrow b_t^j}$ depend only on the previous messages $\mu_{b_t^j\rightarrow a_t^i}$ and vice versa. Thus BP is implemented as a sequential application of $\gv(\cdot)$, $\hv(\cdot)$, $\gv(\cdot)$, $\dots$, and by showing that each of these operations is a contraction, the compound operation $\gv(\hv(\cdot))$ is shown to be a contraction.

\subsection{Bound on complexity}
\label{ss:BPComplexity}
{\noindent}Having proven convergence of the algorithm, we now switch to analysing the computational complexity involved in attaining convergence. We use the term \emph{error} to describe the difference between the exact solution \eqref{eq:MarginalA}, \eqref{eq:MarginalB} and the approximation obtained from BP upon convergence (since BP is an approximate method). In contrast, we use the term \emph{deviation} to describe the difference between the marginal estimate (belief) obtained from BP at a particular iteration and the belief upon convergence. 

We start with a lemma that relates the deviation in the BP messages to the deviation in the corresponding belief. For convenience we define the shorthand $p^i(j)\triangleq \hat{p}(a^i=j|Z^t)$.
\begin{lemma}
If $\left|\log\frac{\nu_{j\rightarrow i}}{\tilde{\nu}_{j\rightarrow i}}\right|\leq\epsilon$ $\forall$ $(i,j)$ and $p$ and $\tilde{p}$ are calculated from $\nu$ and $\tilde{\nu}$ via \eqref{eq:Marginal1}, then $|p^i(j)-\tilde{p}^i(j)|\leq \delta(\epsilon)$ $\forall$ $(i,j)$ where $\delta(\epsilon)=\exp(2\epsilon) - 1\approx 2\epsilon$ (for small $\epsilon$).
\label{lem:MessageErrMarginalErr}
\end{lemma}
\begin{proof}
We seek to bound
\begin{align}
&\left|
\frac{\nu_{j\rightarrow i}\psi_i(j)}{\sum_{j'}\nu_{j'\rightarrow i}\psi_i(j')} - 
\frac{\tilde{\nu}_{j\rightarrow i}\psi_i(j)}{\sum_{j'}\tilde{\nu}_{j'\rightarrow i}\psi_i(j')}
\right| \notag\\
&=
\left|
\frac{\nu_{j\rightarrow i}\psi_i(j)}{\tilde{\nu}_{j\rightarrow i}\psi_i(j)}\cdot
\frac{\sum_{j'}\tilde{\nu}_{j'\rightarrow i}\psi_i(j')}{\sum_{j'}\nu_{j'\rightarrow i}\psi_i(j')} - 1
\right| \cdot
\frac{\tilde{\nu}_{j\rightarrow i}\psi_i(j)}{\sum_{j'}\tilde{\nu}_{j'\rightarrow i}\psi_i(j')} \notag\\
&\leq
\left|
\frac{\nu_{j\rightarrow i}}{\tilde{\nu}_{j\rightarrow i}}\cdot
\frac{\sum_{j'}\tilde{\nu}_{j'\rightarrow i}\psi_i(j')}{\sum_{j'}\nu_{j'\rightarrow i}\psi_i(j')} - 1
\right| \label{eq:MargErrorBound1}
\end{align}
Trivially, $\exp(-\epsilon)\leq\frac{\nu_{j\rightarrow i}}{\tilde{\nu}_{j\rightarrow i}}\leq\exp(\epsilon)$. The second term can be bounded similarly since
\[
\frac{\sum_{j'}\tilde{\nu}_{j'\rightarrow i}\psi_i(j')}{\sum_{j'}\nu_{j'\rightarrow i}\psi_i(j')} 
\leq \frac{\sum_{j'}\exp(\epsilon)\nu_{j'\rightarrow i}\psi_i(j')}{\sum_{j'}\nu_{j'\rightarrow i}\psi_i(j')} \\
= \exp(\epsilon)
\]
Substituting these bounds into \eqref{eq:MargErrorBound1} we find
\[
\exp(-2\epsilon)-1 \leq
\frac{\nu_{j\rightarrow i}}{\tilde{\nu}_{j\rightarrow i}}\cdot
\frac{\sum_{j'}\tilde{\nu}_{j'\rightarrow i}\psi_i(j')}{\sum_{j'}\nu_{j'\rightarrow i}\psi_i(j')} - 1
\leq \exp(2\epsilon)-1
\]
The desired result is then obtained by observing that $0\leq 1-\exp(-2\epsilon)\leq \exp(2\epsilon)-1$. Finally, note that $\exp(2\epsilon)-1 = 2\epsilon + O(\epsilon^2)$ for small $\epsilon$, so that the error is well-approximated as $2\epsilon$.
\end{proof}

Using the previous lemma, we now show that the number of iterations required for the deviation to be less than a desired level is bounded through a simple closed-form expression. In what follows, we denote by $\mu_{i\rightarrow j,k}$ and $\nu_{j\rightarrow i,k}$ the $k$-th iterate of the messages from \eqref{eq:FixedPoint1} and \eqref{eq:FixedPoint2}, and by $\mu_{i\rightarrow j,*}$ and $\nu_{j\rightarrow i,*}$ the messages upon convergence.

\begin{theorem}
\label{th:ClosedFormIterationBound}
Starting from $\nu_{j\rightarrow i,0}=1$ and given $W_*$ and $W^*$ defined in Lemmas \ref{lem:BPFixedPoint1Contraction} and \ref{lem:BPFixedPoint2Contraction}, the deviation between the current beliefs and those obtained at convergence is guaranteed to be no more than $\epsilon$ if the number of iterations $k$ satisfies
\begin{equation}\label{eq:ClosedFormIterationBound}
k-1 \geq \frac{\log[\exp(2\epsilon)-1] - \log\log(1+W_*)}{\log\alpha(1 \! + \! W_*,W_*)+\log\alpha(1 \! + \! W_*,W^*)}
\end{equation}
Looser bounds that only consider the contractions corresponding to one half-message also apply:
\begin{align}
k-1 \geq \frac{\log[\exp(2\epsilon)-1] - \log\log(1+W_*)}{\log\alpha(1 \! + \! W_*,W_*)} \label{eq:LoosenedBound1}\\
k \geq \frac{\log[\exp(2\epsilon)-1] - \log\log(1+W^*)}{\log\alpha(1+W^*,W^*)}
\end{align}
\end{theorem}
\begin{proof}
First, we bound the distance between the messages at initialisation. The point which we choose to bound is the message $\mu_{i\rightarrow j,0}$ which results from a single application of \eqref{eq:FixedPoint1} from the initialisation $\nu_{j\rightarrow i,0}=1$. We examine the ratio
\[
\frac{\mu_{i\rightarrow j,0}}{\mu_{i\rightarrow j,*}} = \frac{1+\sum_{j'\neq j,j'>0}\psi_i(j')\nu_{j'\rightarrow i,*}}{1+\sum_{j'\neq j,j'>0}\psi_i(j')\nu_{j'\rightarrow i,0}}
\]
It is clear that any $\nu_{j\rightarrow i}$ resulting from \eqref{eq:FixedPoint2} will satisfy $0<\nu_{j\rightarrow i}\leq 1$. Consequently,
\[
\frac{1}{1+W_*} \leq \frac{1}{1+W_i} \leq \frac{\mu_{i\rightarrow j,0}}{\mu_{i\rightarrow j,*}} \leq 1
\]
where $W_i$ and $W_*$ were defined in \lemref{lem:BPFixedPoint1Contraction}. Thus, after half an iteration,
\[
\left|\log\frac{\mu_{i\rightarrow j,0}}{\mu_{i\rightarrow j,*}}\right| \leq \log(1+W_*)
\]
After one more half-iteration (neglecting the contraction that this introduces for convenience), we obtain:
\[
\left|\log\frac{\nu_{j\rightarrow i,1}}{\nu_{j\rightarrow i,*}}\right| \leq \log(1+W_*)
\]
Thus we set $\bar{L}=1+W_*$, to obtain a contraction factor $\alpha=\alpha(1 \! + \! W_*,W_*)\alpha(1 \! + \! W_*,W^*)$ (combining the two half-iterations). Then after $(k-1)$ subsequent iterations, the deviation will be bounded by
\[
\left|\log\frac{\nu_{j\rightarrow i,k}}{\nu_{j\rightarrow i,*}}\right| \leq \alpha^{k-1}\log(1+W_*)
\]
In order to obtain the desired bound on the belief deviation, we seek $k$ such that
\begin{align*}
\alpha^{k-1}\log(1+W_*) &\leq \exp(2\epsilon)-1 \\
(k-1) \log{\alpha} &\leq \log[\exp(2\epsilon)-1] - \log\log(1+W_*) \\
k-1 &\geq \frac{\log[\exp(2\epsilon)-1] - \log\log(1+W_*)}{\log{\alpha}}
\end{align*}
This proves the result in \eqref{eq:ClosedFormIterationBound}. The first loosened bound is obtained by observing that $\alpha\leq\alpha(1 \! + \! W_*,W_*)$, \ie, only incorporating the contraction factor for one of the two half-iterations. The second is obtained by observing that
\[
\frac{1}{1+W^*} \leq \nu_{j\rightarrow i,*} \leq 1
\]
and following similar subsequent steps.
\end{proof}

The analysis in \thref{th:ClosedFormIterationBound} can be tightened via a bound that is computed numerically. As iterations progress, we are guaranteed to be closer to the optimal solution. Correspondingly, the contraction factor reduces, hastening convergence. The closed-form analysis does not exploit this fact, rather assuming that the contraction factor remains at its original, worst-case value. The tighter, computable bound is described in \coref{co:ComputableIterationBound}.
\begin{corollary}\label{co:ComputableIterationBound}
Assume that we commence from $\nu_{j\rightarrow i,0}=1$, so that %
\[
\left|\log\frac{\mu_{i\rightarrow j,1}}{\mu_{i\rightarrow j,*}}\right|\leq\log(1+W_*)\triangleq \log L_1 \triangleq l_1
\]
Then for $k>1$
\[
\left|\log\frac{\mu_{i\rightarrow j,k}}{\mu_{i\rightarrow j,*}}\right| \leq l_k
\]
where
\begin{align*}
l_k &= \alpha(L_{k-1},W_*) l_{k-1} \\
L_k &= \exp\{l_k\}
\end{align*}
\end{corollary}

\begin{figure}[tb]
\centering
\includegraphics{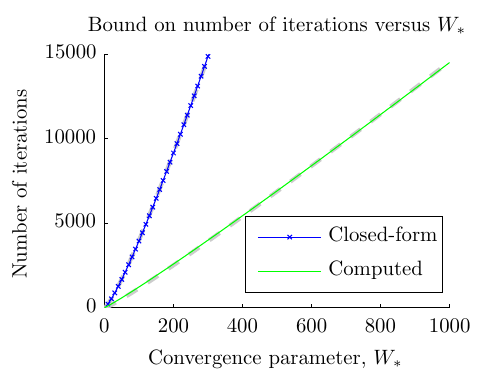}
\caption{Bounds on the number of iterations required for convergence (\ie, to guarantee that the resulting beliefs are within $10^{-3}$ of their final values). The blue line (with $\times$ marks) shows the closed-form bound of \eqref{eq:LoosenedBound1}, while the green line (without symbols) shows the tighter bound computed using \coref{co:ComputableIterationBound}. Least squares fits to the functions $y = c' W_* \log W_*$ (for the closed form bound) and $y = c W_* \log\log W_*$ (for the computable bound) are shown in light-grey dashed lines. \label{fig:Convergence}}
\end{figure}

\figref{fig:Convergence} shows the number of iterations $k$ required to guarantee deviation $\delta=10^{-3}$ in the beliefs for various $W_*$ ranging in $(0,1000]$. The green line shows the computable bound from \coref{co:ComputableIterationBound}, while the blue line (with `$\times$' symbols) shows the closed-form bound from \thref{th:ClosedFormIterationBound} (\ie, the first loosened bound, which depends only on $W_*$). Empirically, the computable bound can be well-approximated by the function $c W_*\log\log W_*$ (for an appropriate choice of $c$), while the closed-form bound can be well-approximated by the function $c' W_*\log W_*$ (for an appropriate choice of $c'$); these functions are shown with light grey dashed lines under the respective bounds.

\subsection{Interpretation of complexity}
\label{ss:BPComplexInterpret}
{\noindent}The previous section showed that the number of iterations required can be bounded by a function that depends on the parameters $W_*=\max_i W_i$ and/or $W^*=\max_j W^j$. In this section, we evaluate the expected value $E_{Z|X}[W_i]$ when the true target positions are $X=\{x^1,\dots,x^n\}$ in order to provide intuition on how this parameter relates to the problem parameters. Modifying notation to explicitly incorporate the measurement set $Z$, we obtain
\[
W_i = \sum_{z\in Z}\psi_i(z)
\]
Simplifying the model to a uniform probability of detection, false alarm density and new target density, the weights $\psi_i(z)$ become
\[
\psi_i(z) = \frac{r_t^i P_d}{\lfan (1-r_t^i P_d)}
\int{f(z|x_t^i)f(x_t^i|Z^{t-1})\dif x_t^i}
\]
The sum in $W_i$ is over all measurements; accordingly its expected value can be calculated using the first moment of the measurement distribution \cite{Mah03,Mah07}
to obtain
\begin{align}
E_{Z|X}[W_i] &= \int{\psi_i(z)\lambda(z|X)\dif z} \\
&= \frac{r_t^i P_d}{\lfan(1-r_t^i P_d)}\int{\lambda(z|X)f^i(z|Z^{t-1})\dif z} \label{eq:ParameterInterpret}
\end{align}
where $f^i(z|Z^{t-1})$ is the distribution of the measurement for the $i$-th track, and $\lambda(z|X)$ is the first moment of the measurement distribution given the true multi-target state X:
\begin{align}
f^i(z|Z^{t-1}) &= \int{f(z|x_t^i)f(x_t^i|Z^{t-1})\dif x_t^i} \\
\lambda(z|X) &= \lfan + \sum_{i=1}^n P_d f(z|x^i)
\end{align}
Accordingly, we interpret \eqref{eq:ParameterInterpret} as being the expected measurement intensity in the vicinity of the predicted measurement distribution for the track $i$. This provides useful intuition on the dependence on various problem parameters:
\begin{itemize}
\item $W_i$ decreases as the false alarm intensity increases. The reduction is more significant when the false alarm intensity is a small contributor to the overall measurement intensity. When the false alarm intensity is the dominant contributor to the intensity, little reduction will occur.
\item $W_i$ decreases as the probability of existence and detection decreases. This occurs both due to the leading factor, and due to the reduction in the overall measurement intensity. As the probability of existence and detection approaches unity, the quantity increases rapidly, due to the term $r_t^i P_d/(1-r_t^i P_d)$ in the leading factor.
\item $W_i$ increases as targets become closely spaced, as multiple targets contribute to the measurement intensity in the region for the track of interest. Highly accurate measurements and dynamic models will reduce $W_i$ if the additional accuracy permits targets to be distinguished, \ie, it reduces the contribution of other targets. If the spacing of the targets is such that the additional accuracy does not separate the targets, an increase in accuracy will increase $W_i$, increasing convergence time.
\end{itemize}
In summary, if additional SNR permits disambiguation of measurement-target association then it will aid convergence; in other cases, lower SNR cases will yield better convergence. Empirically, we will see in \secref{sec:Experiments} that the beliefs also become closer to the exact marginal probabilities in lower SNR conditions. A similar analysis will show that the quantities $W^*$ and $W^j$ depend on the intensity of tracks in the vicinity of measurements, although the quantities involved are not as easily interpreted.

The value of $W_*$ is explored in the experimental results in \figref{fig:ConvParamExperiment} (\secref{ss:ExpResults}).

\subsection{Stopping criterion}
\label{ss:BPStoppingCrit}
{\noindent}The emphasis of the previous sections has been on bounding the number of iterations required to obtain a deviation (\ie, the difference between the estimate at the current iteration and the converged estimate) less than a particular threshold. Practically, this will lead to an algorithm that performs a pre-determined number of iterations. The more common approach in cyclic BP is to compare messages to previous iterates and terminate when convergence is detected. The following theorem, which is a well-known application of contraction mapping results, provides a stopping criterion with a guaranteed bound on the deviation from the converged estimate.

\begin{theorem}
For any $k\geq 2$, if $d(\muv_k,\muv_{k-1}) \leq \epsilon$ then
\[
d(\muv_k,\muv_*) \leq \frac{\alpha(\epsilon,W_*)\alpha(\epsilon,W^*)}{1-\alpha(\epsilon,W_*)\alpha(\epsilon,W^*)} \epsilon
\]
\label{th:ConvergenceTest}
\end{theorem}
\begin{proof}
Let $\muv_{k}$ denote the vector form of the messages at iteration $k$ and $\alpha=\alpha(\epsilon,W_*)\alpha(\epsilon,W^*)$. Then $\forall\;k'\geq k$, $d(\muv_{k'},\muv_{k'-1})\leq \alpha^{k'-k}d(\muv_k,\muv_{k-1}) \leq 
\alpha^{k'-k} \epsilon$ (where the first inequality is by the contraction property, and the second is the assumption in the theorem). By repeated application of the triangle inequality,
\[
d(\muv_k,\muv_*) \leq \sum_{k'=k+1}^\infty d(\muv_{k'},\muv_{k'-1}) \leq \sum_{l=1}^\infty \alpha^l \epsilon = \frac{\alpha}{1-\alpha} \epsilon
\]
\end{proof}
A similar guarantee applies to the alternate messages $\nuv$.

\subsection{Algorithm}
\label{ss:Algorithm}
{\noindent}The BP algorithm is summarised in \figref{fig:Algorithm}. The algorithm incorporates the results of \lemref{lem:MessageErrMarginalErr} and \thref{th:ConvergenceTest} in order to test convergence. Convergence checks are performed every $N$ iterations in order to avoid the computational overhead involved. In practice, we suggest values in the range $N\in\{5,\dots,20\}$.

\begin{figure}
\begin{algorithmic}
\REQUIRE{Number of tracks $n_t$, number of measurements $m_t$, single-target association weights $\psi_i(j)\;\forall\;i\in\{1,\dots,n_t\},\;j\in\{1,\dots,m_t\}$ (assumes $\psi_i(0)=1$), convergence criterion $\delta$, number of iterations between convergence checks $N$}
\ENSURE{Beliefs $\hat{p}_{ij}\;\forall\;i\in\{1,\dots,n_t\},\;j\in\{0,\dots,m_t\}$}
\STATE{$W_* := \max_i \sum_{j>0} \psi_i(j)$} \COMMENT{For convergence criterion}
\STATE{$\nu_{j\rightarrow i} := 1 \;\forall\; i,j>0$}
\REPEAT
\STATE{}\COMMENT{Perform $N$ iterations without checking convergence}
\FOR{$k := 1$ \TO $N$}
  \FOR[Calculate L-R messages]{$i := 1$ \TO $n_t$}
    \STATE{$s := 1 + \sum_{j>0}\psi_i(j)\nu_{j\rightarrow i}$}
    \STATE{$\mu_{i\rightarrow j} := \psi_i(j)/[s - \psi_i(j)\nu_{j\rightarrow i}]\;\forall\;j$}
  \ENDFOR
  \IF[For convergence check]{$k = N$}
  	\STATE{$\tilde{\nu}_{j\rightarrow i} := \nu_{j\rightarrow i}\;\forall\;i,j>0$}
  \ENDIF
  \FOR[Calculate R-L messages]{$j := 1$ \TO $m_t$}
    \STATE{$s := 1 + \sum_{i>0}\mu_{i\rightarrow j}$}
    \STATE{$\nu_{j\rightarrow i} := 1/[s - \mu_{i\rightarrow j}]\;\forall\;i$}
  \ENDFOR
\ENDFOR
\STATE{} \COMMENT{Check for convergence}
\STATE{$d := \max_{i,j}\left|\log\frac{\nu_{j\rightarrow i}}{\tilde{\nu}_{j\rightarrow i}}\right|$}
\STATE{$\alpha := \left(\log\frac{1+W_* d}{1+W_*}\right)/(\log{d})$}
\UNTIL{$\frac{\alpha d}{(1-\alpha)} < \frac{1}{2}\log(1+\delta)$}
\STATE{} \COMMENT{Calculate beliefs}
\FOR{$i:=1$ \TO $n_t$}
\STATE{$s := 1 + \sum_{j>0}\psi_i(j)\nu_{j\rightarrow i}$}
\STATE{$\hat{p}_{i,j} := \psi_i(j)\nu_{j\rightarrow i}/s\;\forall\;j>0$}
\STATE{$\hat{p}_{i,0} := 1/s $}
\ENDFOR
\end{algorithmic}
\caption{Algorithm for computing approximate marginal probabilities (beliefs) using BP.}
\label{fig:Algorithm}
\end{figure}

\section{Relationship to other methods}
\label{sec:Relationships}

\subsection{Junction tree}
\label{ss:JTree}
The \emph{junction tree} algorithm \cite{LauSpi88,KolFri09} is the standard method for conducting exact inference in a cyclic graph. The algorithm provides a systematic procedure for merging\footnote{More precisely, the method forms a tree of hyper-nodes that satisfies the running intersection property, thus ensuring that enforcing consistency along edges is sufficient. For further details, see \cite{KolFri09}.} variables into \emph{hyper-nodes} in order to convert the cyclic graph into a tree, and then executes BP on that tree to conduct exact inference. As previously described, the complexity of the algorithm can be problematic, as the computation increases exponentially in the number of nodes that need to be merged together in order to yield a tree-structured graph.

\begin{figure}[tb]
\centering
\includegraphics{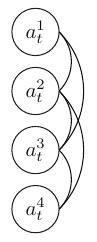}
\caption{Example of graphical formulation used as an input to the junction tree algorithm. In this case, there is no measurement with non-zero weight for both target $1$ and target $4$, hence there is no edge between the corresponding nodes.}\label{fig:JTreeGraph}
\end{figure}

Junction trees can be applied to data association using the graph described in \figref{fig:Bipartite}, but this would be quite inefficient, since the redundant use of target association variables and measurement association variables is of no benefit in the exact case, and it increases the number of variables over which inference must be conducted. Instead, we use the simpler, yet equivalent representation of \eqref{eq:JointDistribution}:
\begin{equation}
p(a_t^1,\dots,a_t^{n_t}|Z^t) \propto \left[\prod_{i=1}^{n_t}\psi_i(a^i)\right] \cdot
\left[\prod_{(i,i')\in\calE}\psi_c(a^i,a^{i'})\right]
\end{equation}
where
\begin{align}
\psi_c(a^i,a^{i'}) &= \begin{cases}
0, & a^i=a^{i'} > 0 \\
1, & \mbox{otherwise}
\end{cases} \\
\calE &= \big\{
(i,i') \big| \exists \;i\in\{1,\dots,n_t\},i'\in\{1,\dots,n_t\},\notag\\
&\qquad j\in\{1,\dots,m_t\} \mbox{ s.t.\ } i\neq i',\;\psi_i(j)\psi_{i'}(j)>0
\big\}\label{eq:JTEdgeSet}
\end{align}
We assume that the node potentials $\psi_i(j)$ have been thresholded such that they are zero for infeasible associations (\eg, due to gating). This in turn creates the sparsity in $\calE$ which is exploited by the junction tree algorithm. While BP could also be applied directly to the cyclic graph in this formulation (as proposed in \cite{AngSri04}, \cite[Box 12.D]{KolFri09}), it was shown in \cite{WilLau10} that it performs poorly (in both accuracy and convergence) in comparison to the bipartite formulation in \eqref{eq:JointDistribution}. An example of this graph is show in \figref{fig:JTreeGraph}.

The efficient hypothesis management method (EHM-2) \cite{MasBri04,HorMas06} exploits a similar tree-based inference, gaining additional efficiency by effectively reducing the alphabet within the junction tree hyper-nodes (\ie, eliminating elements of the hyper-node alphabet which violate mutual exclusion constraints and thus have zero probability).

The experiments in \secref{sec:Experiments} compare the accuracy and computation time of the method proposed in this paper to the junction tree algorithm with various thresholds applied to $\psi_i(j)$, yielding various levels of sparsity in the resulting graph. The experiments were conducted using the implementation of junction tree algorithm contained in the library for discrete approximate inference \cite{libDAI}.

\subsection{Correlation decay}
\label{ss:CorrDecay}
{\noindent}The recently proposed method of \cite{Oh11} utilises the correlation decay property of statistical physics to obtain a deterministic FPTAS for the marginal association probabilities. The algorithm involves a recursion of the form
\begin{align}
\Phi(\calT,\calM,i,t) &= \frac{1}{1 + \sum_{j\in\calM}\psi_i(j)\Phi(\calT-\{i\},\calM,j,t-1)} \label{eq:CorrDecay1}\\
\Phi(\calT,\calM,j,t) &= \frac{1}{1 + \sum_{i\in\calT}\psi_i(j)\Phi(\calT,\calM-\{j\},i,t-1)} \label{eq:CorrDecay2}
\end{align}
where $\calT$ (respectively, $\calM$) is the set of tracks (respectively, measurements) remaining in the recursion, and $t$ is the maximum number of recursion steps to perform. Close inspection reveals that these equations are almost identical to \eqref{eq:FixedPoint1} and \eqref{eq:FixedPoint2}; the difference is that the recursion in \eqref{eq:CorrDecay1} and \eqref{eq:CorrDecay2} never revisits nodes in its recursion; consequently it is exact, but has exponential complexity. 

The theoretical analysis in \cite{Oh11} shows that if the number of nodes in the graph grows, but the maximum connectivity and maximum single-target association weight remains constant, $t$ may be chosen such that the error is bounded yet complexity is polynomial in the number of tracks and measurements. The proof of this theorem, which is based on the earlier work \cite{BayGam07}, exploits similar properties to the convergence proof in \secref{ss:BPConvergence}.

The experiments in \secref{sec:Experiments} compare the accuracy and computation time of the correlation decay algorithm to the method proposed in this paper.

\subsection{Weighted Bethe energy}
\label{ss:WeightedBethe}
As discussed in \secref{ss:BackMatching}, the present work is closely related to \cite{Von13}. A recent extension of this work in \cite{CheYed13} studies the effect of applying a weight $\beta\in[0,1]$ to a subset of the terms in the Bethe free energy (in the form derived in \cite{Von13}). The work is particularly interesting in the context of estimating permanents, since there is guaranteed to be a $\beta\in[0,1]$ which yields the exact answer. Convergence of BP in this modified objective remains to be proven. Extending the convergence proof to this case and examining whether it may provide improved estimates of marginal probabilities are both topics of future study.

%%%%%%%%%%%%%%%%%%%%%%%%%%%%%%%%%%%%%%%%%%%%%%%%%%%%%%%%%%%%%%%%%%%%%%%%%%%%%%%
\section{Experiments}
\label{sec:Experiments}
{\noindent}We consider a series of single time-step problems involving targets on a regularly spaced grid. Although the experiment involves only a single time step, the track covariances are preinitialised by simulating $30$ time steps of the standard constant velocity model:
\begin{align*}
P_{t|t-1} &= F P_{t-1|t-1} F^T + Q \\
P_{t|t} &= \begin{cases}
P_{t|t-1}, & \mbox{target not detected} \\
P_{t|t-1} - K H P_{t|t-1}, & \mbox{target detected}
\end{cases} \\
K &= P_{t|t-1} H^T (H P_{t|t-1} H^T + R)^{-1}
\end{align*}
with
\[
F = I_{2\times 2} \otimes \left[\begin{array}{cc}
1 & T \\
0 & 1
\end{array} \right], \quad
Q = q I_{2\times 2} \otimes \left[\begin{array}{cc}
T^3/3 & T^2/2 \\
T^2/2 & T
\end{array} \right],
\]
$T=1$, $q=0.01$, $H = I_{2\times 2} \otimes \left[\begin{array}{cc} 1 & 0 \end{array} \right]$, and $R = r I_{2\times 2}$, commencing with $P_{0|0}$ equal to zero. The track estimates were initialised to the true target positions, corrupted with additive Gaussian noise with covariance distributed according to the prior covariance for the target, generated through independent simulation of each target. Gating was performed with a threshold such that the probability of excluding the target-derived measurement was $10^{-4}$. The area populated with false alarms was sufficiently large to cover these gates.

\begin{figure*}[!t]
\centering
\includegraphics{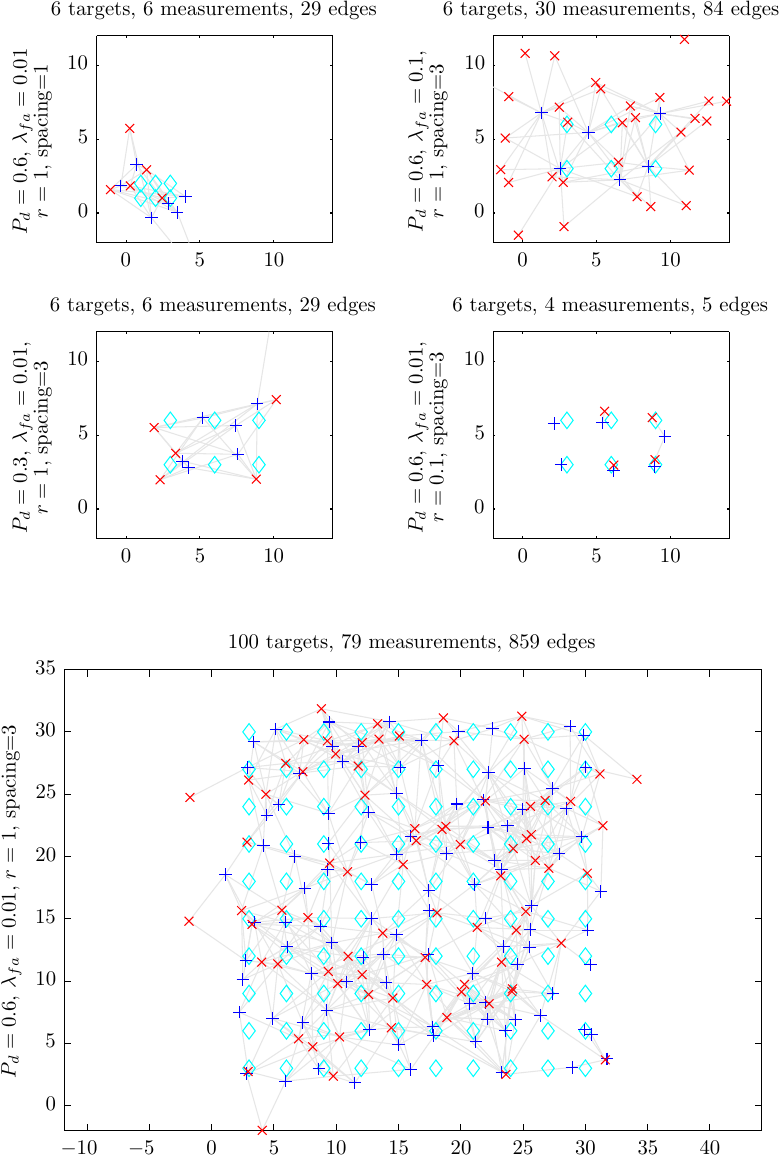}
\caption{Examples of single Monte Carlo trials. True target positions are shown as {\color[rgb]{0,1,1}$\Diamond$}, measurements (false alarms and true measurements) are shown as {\color[rgb]{1,0,0}$\times$}, and prior track estimates are shown as {\color[rgb]{0,0,1}$+$}. Faint grey lines show the edges in the graph, connecting track estimates and measurements which form feasible associations.}
\label{fig:Examples}
\end{figure*}

Three groups of experiments were conducted:
\begin{enumerate}
\item Six targets (arranged in a regular $2\times 3$ grid), varying target spacing between $0$ and $10$ units, considering cases with $P_d\in\{0.3,0.6,0.9\}$, $\lfa\in\{0.01,0.0316,0.1\}$, and $r\in\{0.1,1,10\}$
\item Targets in a $n\times 3$ regular grid (with $n\in\{2,\dots,30\}$) with target spacing set to $3$ units, $P_d=0.6$, $\lfa=0.01$ and $r=1$
\item Targets in a $n\times n$ regular grid (with $n\in\{2,\dots,10\}$) with target spacing set to $3$ units, $P_d=0.6$, $\lfa=0.01$ and $r=1$
\end{enumerate}
A number of example experiments are shown in \figref{fig:Examples}.

For each parameter (spacing, $P_d$, \etc.), $1000$ single time step Monte Carlo trials were performed. The total number of Monte Carlo trials performed for each algorithm was $395,000$. The experiments were performed using a dual processor Intel Xeon E5-2670 server with Matlab Parallel Computing Toolbox, utilising $12$ worker threads.

To evaluate the various algorithms compared, we calculated the average maximum error in the marginal probability estimates calculated by each algorithm, and the average computation time. The comparison is performed on the basis of the accuracy of the marginal estimates in order to exclude down-stream effects such as coalescence, mixture reduction, \etc. While these effects are significant, they are separate issues to the approximation of the marginal probabilities, and any conclusion reached in an experiment including these effects is specific to the full gamut of approximations made in the tracking system. We define the average maximum error as being the average value of the largest element of the difference between the marginal distribution of a target association variable estimated by the algorithm under test, and the corresponding reference value, thus averaging is performed over both Monte Carlo trials and over targets in those trials. In the first experiment, the reference value is the exact marginal distribution, calculated using the junction tree algorithm with a threshold of $10^{-4}$ (which is applied to the weights utilised in all algorithms). The size of the problems in the second and third experiments prevented exact computation, thus the reference values were calculated using MCMCDA with $10^7$ samples. A second instance of this algorithm was included as a test, providing an estimate of the expected error in the reference values.

\subsection{Comparison algorithms}
\label{ss:ExpAlgorithms}
{\noindent}The algorithms compared in the experiments were:
\begin{itemize}
\item The BP algorithm detailed in \figref{fig:Algorithm}. A vectorised, Matlab-based implementation was utilised.
\item The junction tree algorithm described in \secref{ss:JTree}, with weights thresholded to $10^{-3}$, $10^{-2}$ and $10^{-1}$, inducing varying degrees of sparsity, and thus exploring the accuracy versus computation time trade-off. The algorithm was implemented using the general-purpose libDAI system \cite{libDAI}.
\item The MCMCDA algorithm of \cite{OhRus09}, using $10^5$, $10^6$ and $10^7$ MCMC steps. The algorithm was initialised with the MAP association (calculated via an auction) to avoid the need for burn-in. The implementation was written in Matlab and compiled into C++ using the Matlab Coder.
\item The correlation decay algorithm described in \secref{ss:CorrDecay}, using $t\in\{3,5,7\}$. The implementation was written in C++.
\item The linear multitarget integrated existence PDA (LMIPDA) algorithm of \cite{MusLaS08}. The track existence probabilities were set to unity for comparison to non-IPDA methods. A vectorised, Matlab-based implementation was utilised.
\item The approximate Bakhtiar-Alavi-Amoozegar (BAA) algorithm from \cite{BakAla96}, which was shown to be the approximation of choice in \cite{RomCro10}. A vectorised, Matlab-based implementation was utilised. 
\end{itemize}
The preliminary version of this paper \cite{WilLau10} included results for PDA (\ie, ignoring the presence of adjacent targets), and BP applied directly to the cyclic graph in the alternative formulation in \figref{fig:JTreeGraph} (as discussed in \secref{ss:JTree}). These were excluded from this comparison as the errors they committed were considerably worse than the algorithms we are comparing. 

As discussed in \secref{ss:JTree}, EHM-2 \cite{MasBri04,HorMas06} exploits problem-specific structure to provide a more efficient version of the junction tree algorithm. Comparison to EHM-2 is the topic of future work. Since the approach is exact, the results would be equivalent to the junction tree algorithm; the point of interest would be the computational efficiency.

\ifCLASSOPTIONdraftcls
\begin{figure*}[!t]
\centering
\includegraphics[width=6in]{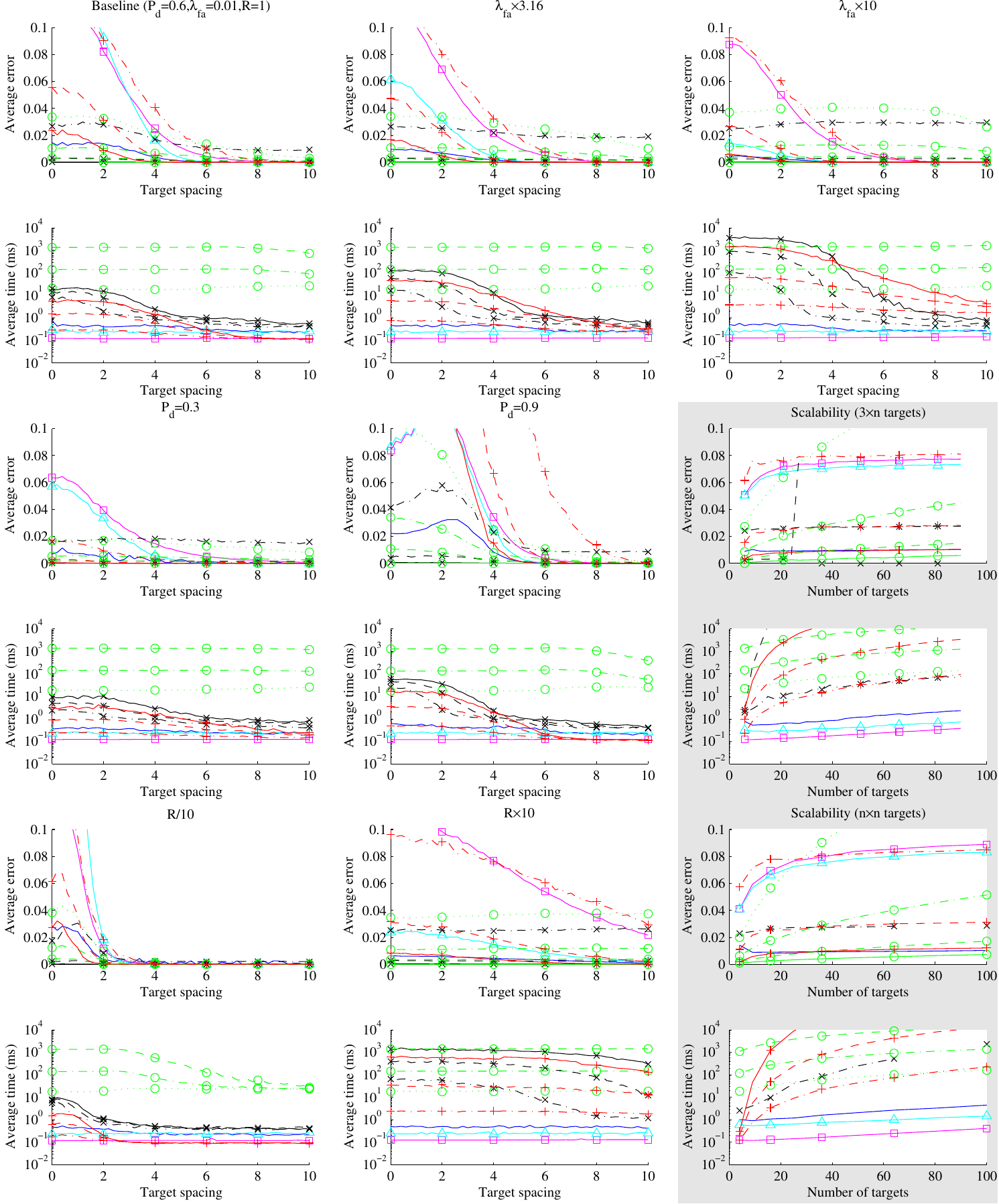}
\caption{Results of experiments. First, third and fifth rows of plots show the average worst-case error in the marginal probabilities, averaged over targets and Monte Carlo trials. Second, fourth and sixth rows show the average computation time. Unshaded plots show results of the first experiment (involving six targets on a regular grid with spacing varied on the $x$-axis), while shaded plots show the second and third experiments (with $3\times n$ and $n\times n$ targets and respectively). Colours and/or symbols show algorithms as {\color[rgb]{0,0,1}BP} (plain lines),  {\color[rgb]{1,0,1}LMIPDA} (triangles), {\color[rgb]{0,1,1}BAA} (squares), {\color[rgb]{0,1,0}MCMCDA} (circles, with dot-dashed, dashed and solid representing $10^5$, $10^6$ and $10^7$ iterations), {\color[rgb]{1,0,0}correlation decay} (`$+$' symbols, with dot-dashed, dashed and solid representing $3$, $5$ and $7$ iterations) and junction tree (`$\times$' symbols, with dot-dashed, dashed and solid representing thresholds of $10^{-1}$, $10^{-2}$ and $10^{-3}$). 
}\label{fig:Results}
\end{figure*}
\else
\begin{figure*}[!t]
\centering
\includegraphics[width=7in]{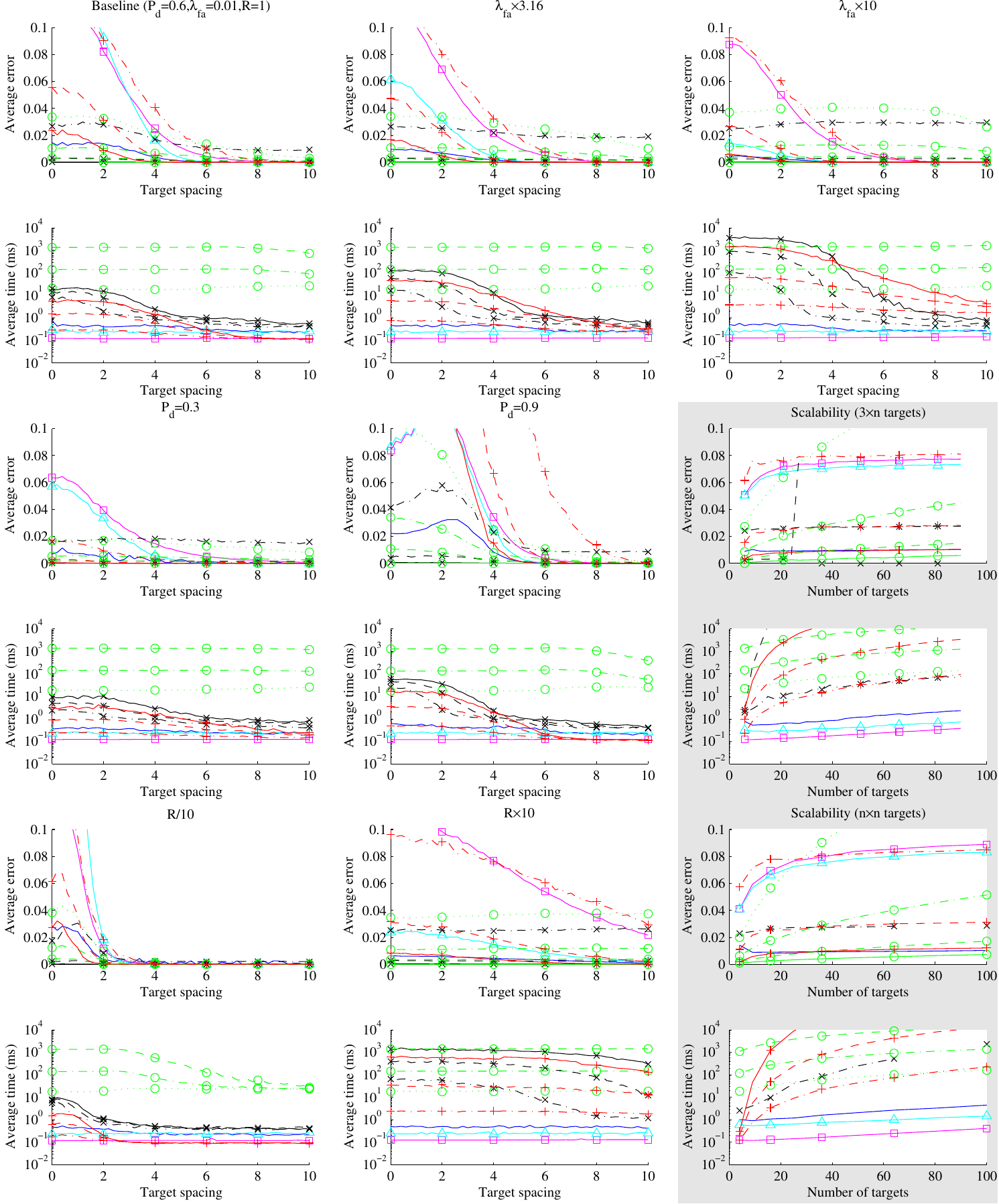}
\caption{Results of experiments. First, third and fifth rows of plots show the average worst-case error in the marginal probabilities, averaged over targets and Monte Carlo trials. Second, fourth and sixth rows show the average computation time. Unshaded plots show results of the first experiment (involving six targets on a regular grid with spacing varied on the $x$-axis), while shaded plots show the second and third experiments (with $3\times n$ and $n\times n$ targets and respectively). Colours and/or symbols show algorithms as {\color[rgb]{0,0,1}BP} (plain lines),  {\color[rgb]{1,0,1}LMIPDA} (triangles), {\color[rgb]{0,1,1}BAA} (squares), {\color[rgb]{0,1,0}MCMCDA} (circles, with dot-dashed, dashed and solid representing $10^5$, $10^6$ and $10^7$ iterations), {\color[rgb]{1,0,0}correlation decay} (`$+$' symbols, with dot-dashed, dashed and solid representing $3$, $5$ and $7$ iterations) and junction tree (`$\times$' symbols, with dot-dashed, dashed and solid representing thresholds of $10^{-1}$, $10^{-2}$ and $10^{-3}$). 
}\label{fig:Results}
\end{figure*}
\fi

\subsection{Results and discussion}
\label{ss:ExpResults}
{\noindent}The results of the comparison are shown in \figref{fig:Results}. The rows of plots alternate between average marginal error, and average computation time (per simulation). Plots with shaded backgrounds correspond to the second and third experiments (\ie, varying numbers of targets), while the remaining plots correspond to the six target experiments. The $x$-axis in the six-target experiments shows the spacing of the true positions of the targets in the regular grid, varying between $0$ and $10$ units. The first two rows of plots show results varying the false alarm rate with $\lfa\in\{0.01,0.0316,0.1\}$, with $P_d = 0.6$ and $r=1$. The first case $\lfa=0.01$ is the baseline, to which the following cases compare. The second two rows show the effect of lowering or raising the $P_d$, \ie, $P_d\in\{0.3,0.9\}$. The final two rows show the effect of raising or lowering the measurement noise, \ie, $r\in\{0.1,10\}$. The top two plots in the shaded box show the second experiment (with $3\times n$ targets varying $n\in\{2,\dots,30\}$, where the $x$-axis shows the total number of targets), and the third experiment (with $n\times n$ targets varying $n\in\{2,\dots,10\}$, again with the $x$-axis showing the total number of targets). In each of these latter cases, the target spacing is fixed to $3$ units, and baseline parameters apply.

We make the following observations:
\begin{itemize}
\item BP (blue) exhibits excellent performance, with average errors of $0.015$ or less in most cases, except for the high $P_d$ case ($P_d=0.9$) and the low measurement noise case ($r=0.1$); in these two cases, the errors are in the $0.02-0.04$ range. Note that these two cases correspond to high SNR. The large scale problems ($3\times n$ and $n\times n$) show little or no indication that the accuracy of BP deteriorates as the number of targets increases.
\item BP (blue) significantly outperforms LMIPDA (magenta) and BAA (cyan), exhibiting errors reduced by a factor of ten in most cases. The computational complexity of BP is only a few times that of LMIPDA and BAA, averaging fractions of a millisecond in all of the six-target problems, and a little over a millisecond even in the $90$ and $100$ target problems.
\item LMIPDA (magenta) and BAA (cyan) appear to perform similarly to each other across the board. BAA performs significantly better in high false alarm and high measurement noise cases, while LMIPDA performs better in low measurement noise cases. The computation time is comparable. 
\item To reiterate, although junction tree is an exact algorithm, we apply it to different approximations of the problem (thresholding the weights) to obtain a trade-off between accuracy and computation. The method (shown in black) exhibits excellent performance with smaller thresholds, but at a high computational cost. With the threshold set to $0.1$ (black dash-dotted), the error is globally worse than BP, and often much worse. The computational complexity in this case is $1$-$2$ orders of magnitude higher than BP, and the $9\times 9$ and $10\times 10$ experiments were unable to be completed due to the high memory requirements of the algorithm. With the threshold set to $0.01$ or $0.001$, the accuracy of junction tree is generally much better than BP, but the computation time is $2$-$4$ orders of magnitude higher than BP. Most experiments in the $3\times n$ and $n\times n$ cases were unable to be completed due to excessive memory requirements.
\item The performance of MCMCDA (green) with $10^5$ iterations (dot-dashed) is generally comparable to BP in the six-target cases, but it performs progressively worse in cases with larger number of targets. In the larger cases, BP performs similar to or better than MCMCDA with $10^6$ iterations. The complexity of MCMCDA is $2$-$4$ orders of magnitude higher than BP depending on the number of samples used.
\item BP generally outperforms the correlation decay method (red) with $t=3$ (dot-dashed) and $t=5$ (dashed). With $t=7$, BP generally performs better for tighter target spacings, and correlation decay performs better with larger target spacings. The correlation decay method performs particularly poorly in the case with $P_d=0.9$. The computational complexity of the correlation decay method varies greatly through the experiments. In some cases (\eg, $r=0.1$) it is slightly faster than BP. In many other cases its complexity is $2$-$3$ orders of magnitude slower, and in the large scale problems, it is $4$-$5$ orders of magnitude slower.
\end{itemize}

\begin{figure}[tb]
\centering
\includegraphics{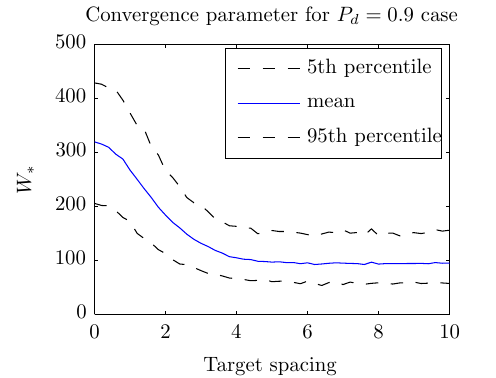}
\caption{Convergence parameter $W_*$ as a function of target spacing for the experiment with $P_d=0.9$. The solid blue line shows the mean value, while the dashed black lines show the 5th and 95th percentiles (over the 1000 experiments).
}\label{fig:ConvParamExperiment}
\end{figure}

The convergence parameter $W_*$ is analysed in \figref{fig:ConvParamExperiment} for the experiment with the most challenging convergence, namely with $P_d=0.9$. The figure illustrates that the convergence parameter is related to the density of measurements in the vicinity of tracks. It is minimal when targets are well-spaced, and maximal when targets are closely-spaced. Since the convergence parameter is around 300, \figref{fig:Convergence} shows a bound on the number of iterations of around 3000. In practice, the maximum number of iterations in the experiment is 87; this difference shows that the offline bounds (both closed-form and computable) examined in \figref{fig:Convergence} are very loose compared to the online termination criterion provided by \thref{th:ConvergenceTest}.

\section{Conclusion}
\label{sec:Conclusion}
This paper has introduced a new approximate method for solving data association problems using BP on a particular graphical model formulation. While BP is generally not guaranteed to converge on cyclic graphs, we have proven convergence on the formulation studied, and bounded the computation required. While the method is approximate, the experiments in \secref{sec:Experiments} reveal a highly favourable comparison with state-of-the-art methods in the accuracy versus computation time trade-off. Future work includes extensions to problems involving multiple time steps or multiple sensors, extensions to obtain guarantees on the accuracy of the approximation, evaluation of the impact to accuracy and convergence of the weighting proposed in \cite{CheYed13}, and comparison with the computational complexity of EHM \cite{HorMas06}.

%%%%%%%%%%%%%%%%%%%%%%%%%%%%%%%%%%%%%%%%%%%%%%%%%%%%%%%%%%%%%%%%%%%%%%%%%%%%%%%
\section*{Acknowledgements}
{\noindent}The authors thank Prof.\ Alan Willsky and Prof.\ M\"{u}jdat \c{C}etin for discussions during the development of the formulation in \secref{sec:BPDA}, Dr Stephen Howard for discussions in the early stage of the convergence proof, and the anonymous reviewers for suggestions that helped to clarify many points. 

\ifCLASSOPTIONdraftcls
\else
\IEEEtriggercmd{\enlargethispage{-1.2in}}
\IEEEtriggeratref{30}
\fi

{\small{\bibliographystyle{IEEEtran}
\bibliography{IEEEabrv,../Bibliography}}}

\begin{IEEEbiography}[{\includegraphics[width=1in,height=1.25in,clip,keepaspectratio]{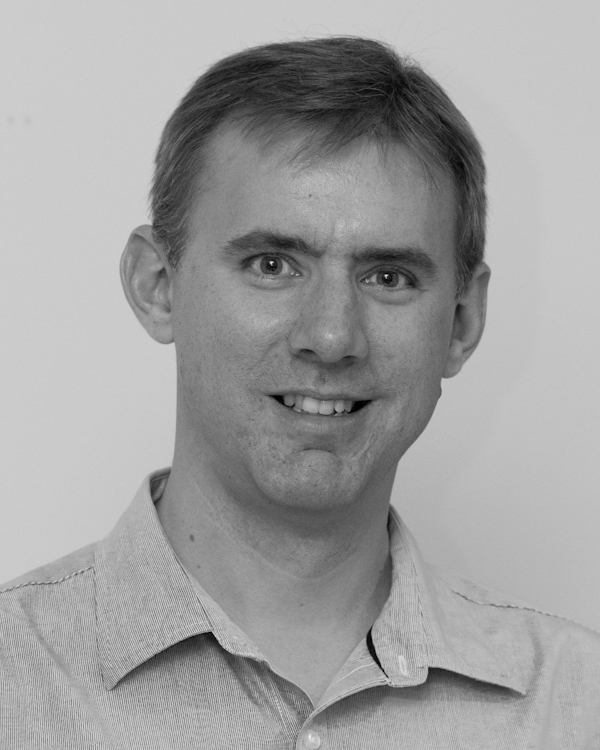}}]{Jason L.\ Williams} (S'01--M'07) received degrees of BE(Electronics)/BInfTech from Queensland University of Technology in 1999, MSEE from the United States Air Force Institute of Technology in 2003, and PhD from Massachusetts Institute of Technology in 2007. 

He worked for several years as an engineering officer in the Royal Australian Air Force, before joining Australia's Defence Science and Technology Organisation in 2007. He is also an adjunct senior lecturer at the University of Adelaide. His research interests include target tracking, sensor resource management, Markov random fields and convex optimisation. 
\end{IEEEbiography}

\begin{IEEEbiography}[{\includegraphics[width=1in,height=1.25in,clip,keepaspectratio]{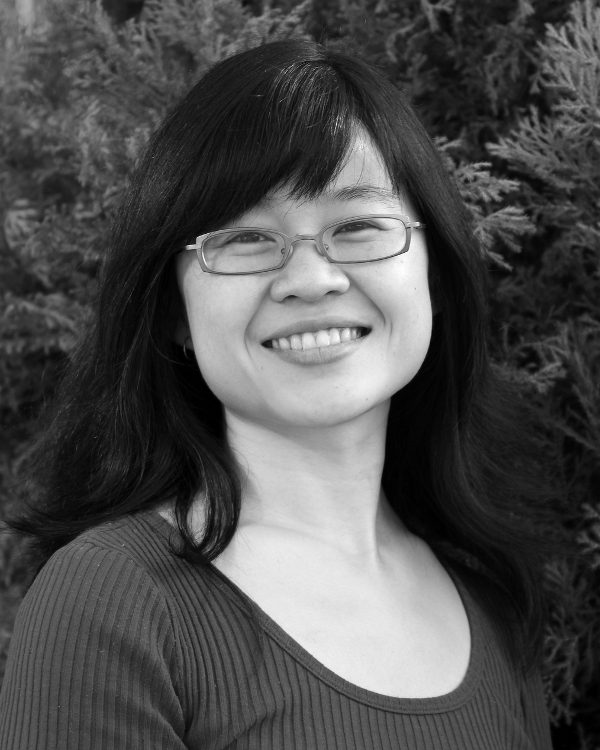}}]{Roslyn A.\ Lau} (S'14) received the degrees of BE(Computer Systems)/BMa\&CS(Statistics) in 2005, and MS(Signal Processing) in 2009, all from the University of Adelaide, Australia. She is currently a PhD candidate at the Australian National University. She is also a scientist at the Defence Science and Technology Organisation, Australia. Her research interests include target tracking, data fusion and graphical models.
\end{IEEEbiography}

\ifCLASSOPTIONdraftcls
\else
\vfill
\fi

\end{document}